\documentclass{article}
\usepackage[final,nonatbib]{neurips_2020}

\usepackage[utf8]{inputenc} % allow utf-8 input
\usepackage[T1]{fontenc}    % use 8-bit T1 fonts
\usepackage{hyperref}       % hyperlinks
\usepackage{url}            % simple URL typesetting
\usepackage{booktabs}       % professional-quality tables
\usepackage{amsfonts}       % blackboard math symbols
\usepackage{nicefrac}       % compact symbols for 1/2, etc.
\usepackage{microtype}      % microtypography
\usepackage{graphicx}
\usepackage{amsmath}
\usepackage{mathtools}
\usepackage{makecell}
\usepackage{amsthm}
\usepackage{enumitem}
\usepackage{subcaption}
\usepackage[table]{xcolor}
\usepackage{booktabs}
\usepackage{ragged2e}
\usepackage{arydshln}
\usepackage[hang,flushmargin]{footmisc}
\usepackage{multirow}
\usepackage{multicol}

\newtheorem{definition}{Definition}

\newcommand{\smalltimes}{{\mkern-1mu\times\mkern-1mu}}
\newcommand{\norm}[1]{\left\lVert#1\right\rVert}
\usepackage{listings}

\usepackage{color}

\title{
Structured Convolutions for Efficient Neural Network Design}

\author{
Yash Bhalgat 
\And
Yizhe Zhang 
\And 
Jamie Menjay Lin
\And 
Fatih Porikli\\
\and
Qualcomm AI Research\thanks{Qualcomm AI Research is an initiative of Qualcomm Technologies, Inc.}\\
{\tt\small \{ybhalgat, yizhez, jmlin, fporikli\}@qti.qualcomm.com}\\
}

\begin{document}

\maketitle

\begin{abstract}
In this work, we tackle model efficiency by exploiting redundancy in the \textit{implicit structure} of the building blocks of convolutional neural networks. We start our analysis by introducing a general definition of Composite Kernel structures that enable the execution of convolution operations in the form of efficient, scaled, sum-pooling components. As its special case, we propose \textit{Structured Convolutions} and show that these allow decomposition of the convolution operation into a sum-pooling operation followed by a convolution with significantly lower complexity and fewer weights. We show how this decomposition can be applied to 2D and 3D kernels as well as the fully-connected layers. Furthermore, we present a Structural Regularization loss that promotes neural network layers to leverage on this desired structure in a way that, after training, they can be decomposed with negligible performance loss. By applying our method to a wide range of CNN architectures, we demonstrate `structured' versions of the ResNets that are up to 2$\times$ smaller and a new Structured-MobileNetV2 that is more efficient while staying within an accuracy loss of 1\% on ImageNet and CIFAR-10 datasets. We also show similar structured versions of EfficientNet on ImageNet and HRNet architecture for semantic segmentation on the Cityscapes dataset. Our method performs equally well or superior in terms of the complexity reduction in comparison to the existing tensor decomposition and channel pruning methods.

\end{abstract}

\section{Introduction}
Deep neural networks deliver outstanding performance across a variety of use-cases but quite often fail to meet the computational budget requirements of mainstream devices. Hence, model efficiency plays a key role in bridging deep learning research into practice. 
% One strategy to attain efficiency is to compress models. 
Various model compression techniques rely on a key assumption that the deep networks are over-parameterized, meaning that a significant proportion of the parameters are redundant. This redundancy can appear either explicitly or implicitly. In the former case, several structured \cite{he2017,lipruning}, as well as unstructured \cite{han2015deep,han2015pruning, automatedpruningmanessi,admm}, pruning methods have been proposed to systematically remove redundant components in the network and improve run-time efficiency. On the other hand, tensor-decomposition methods based on singular values of the weight tensors, such as spatial SVD or weight SVD, remove somewhat implicit elements of the weight tensor to construct low-rank decompositions for efficient inference \cite{denton2014,jaderberg2014speeding,kuzmin2019taxonomy}.	 
%Deep neural networks deliver excellent performance across a variety of use-cases but quite often fail to meet the computational budget requirements of day-to-day devices. Hence, model efficiency plays a key role in today's deep learning research and practice. Various model compression techniques rely on a key assumption that the deep networks are over-parametrized, meaning that a significant proportion of the deep network's parameters are redundant. This redundancy can appear either explicitly or implicitly. In the former case, several structured as well as unstructured pruning methods have been proposed that systematically remove redundant components in the deep network to improve run-time efficiency \cite{han2015deep,han2015pruning,he2017,lipruning, automatedpruningmanessi,admm}. On the other hand, tensor-decomposition methods based on singular values of the weight tensors, such as Spatial SVD or Weight SVD, remove somewhat implicit elements of the weight tensor to create low-rank decompositions for efficient inference \cite{denton2014,jaderberg2014speeding,kuzmin2019taxonomy}.

Redundancy in deep networks can also be seen as network weights possessing an unnecessarily high degrees of freedom (DOF). Alongside various regularization methods \cite{weightdecay, dropout} that impose constraints to avoid overfitting, another approach for reducing the DOF is by decreasing the number of \textit{learnable} parameters. To this end, \cite{jaderberg2014speeding, dcfnet, basisconv} propose using certain basis representations for weight tensors. In these methods, the basis vectors are fixed and only their coefficients are learnable. Thus, by using a smaller number of coefficients than the size of weight tensors, the DOF can be effectively restricted. But, note that, this is useful only during training since the original higher number of parameters are used during inference. \cite{dcfnet} shows that systematically choosing the basis (e.g. the Fourier-Bessel basis) can lead to model size shrinkage and flops reduction even during inference.
%Redundancy in deep networks can also be seen as the network weights possessing unnecessary degrees of freedom (DOF). Alongside various regularization methods \cite{weightdecay, dropout} that constrain the DOF of network weights to avoid overfitting, another way of reducing their DOF is by reducing the number of \textit{learnable} parameters. \cite{jaderberg2014speeding, dcfnet, basisconv} propose using certain basis representations for the weight tensors. In these methods, the basis vectors are fixed and only their coefficients are learnable. Hence, by using lesser coefficients than the number of weight parameters in the tensors, the DOF can be restricted. But note that this is useful only during training, since the actual higher number of parameters are used during inference. \cite{dcfnet} shows that systematically choosing the basis (e.g. the Fourier-Bessel basis) can lead to model size reduction and flops reduction even during inference.

\begin{figure}
    \centering
    \begin{minipage}{0.3\textwidth}
        \centering
        \includegraphics[width=0.98\textwidth]{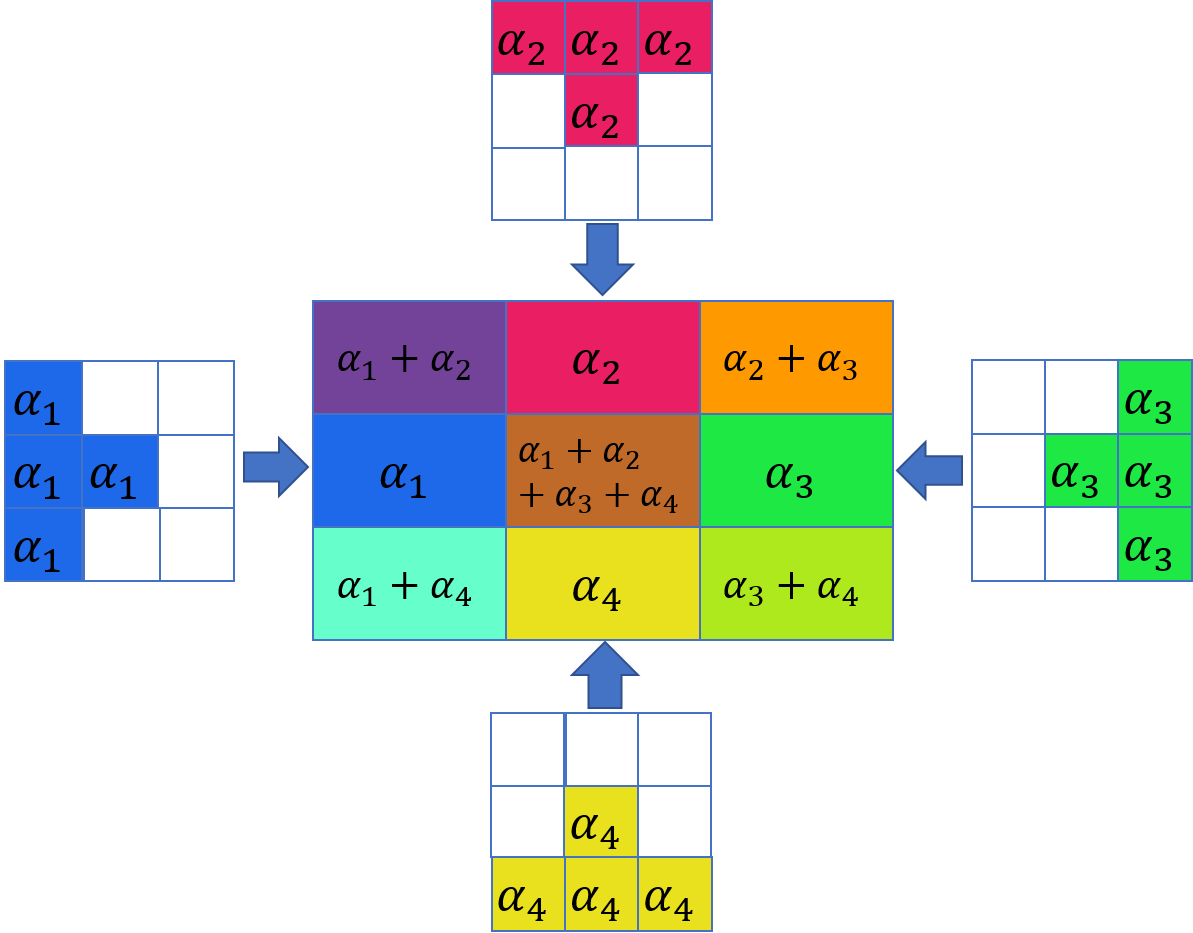}
        (a)
        % first figure itself

    \end{minipage}\hfill
    \begin{minipage}{0.32\textwidth}
        \centering
        \includegraphics[width=0.98\textwidth]{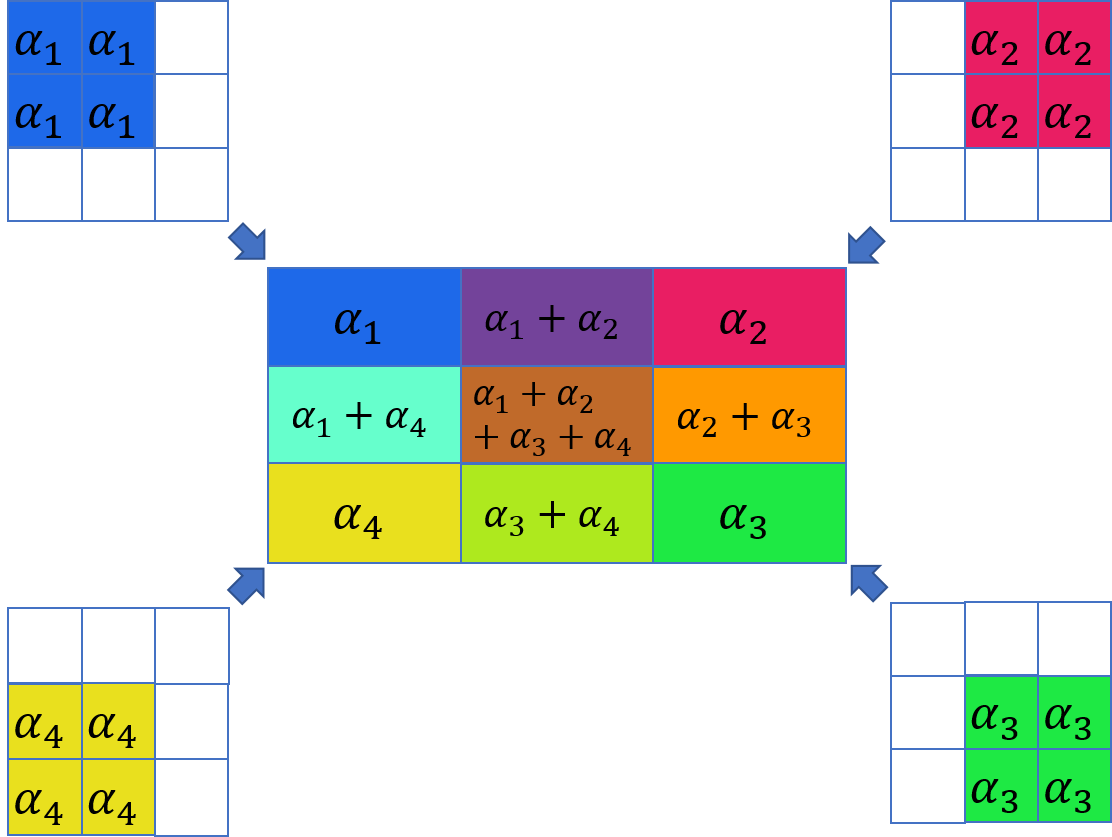}
        (b)
        % first figure itself
    \end{minipage}\hfill
    \begin{minipage}{0.25\textwidth}
        \centering
        \includegraphics[width=0.98\textwidth]{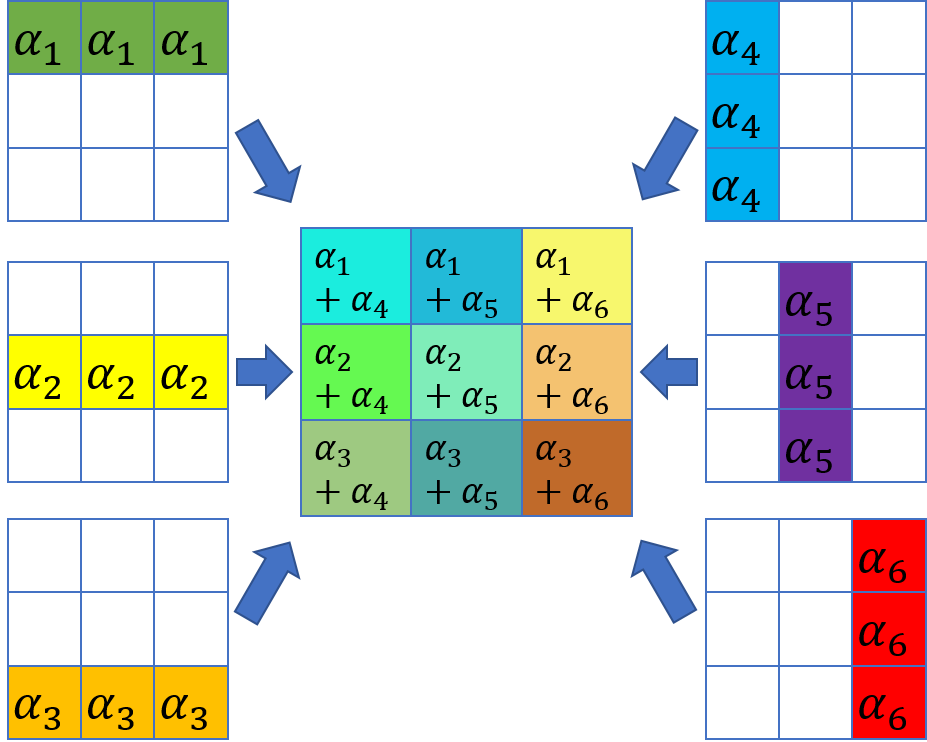}
        (c)
        % first figure itself
    \end{minipage}
    
    \caption{A $3\times 3$ composite kernel constructed as a superimposition of different underlying structures. Kernels in (a) and (b) possess 4 degrees of freedom whereas the kernel in (c) has 6 degrees of freedom. \newline
    \textit{Color combinations are chosen to reflect the summations, this figure is best viewed in color.}}
    \label{fig:OS_illustration}
\end{figure}

In this work, we explore restricting the degrees of freedom of convolutional kernels by imposing a structure on them. This structure can be thought of as constructing the convolutional kernel by super-imposing several \textit{constant-height} kernels. A few examples are shown in Fig.~\ref{fig:OS_illustration}, where a kernel is constructed via superimposition of $M$ linearly independent masks with associated constant scalars $\alpha_m$, hence leading to $M$ degrees of freedom for the kernel. 
% The binary nature of these masks enables efficient execution of the convolution operation as explained in Sec. \ref{sec:conv_composite}. \yb{I put back the older sentence because I think it gives a slightly positive tone}
The very nature of the basis elements as binary masks enables efficient execution of the convolution operation as explained in Sec. \ref{sec:conv_composite}.

In Sec.~\ref{sec:structuredconv}, we introduce \textit{Structured Convolutions} as a special case of this superimposition and show that it leads to a decomposition of the convolution operation into a sum-pooling operation and a significantly smaller convolution operation. We show how this decomposition can be applied to convolutional layers as well as fully connected layers. We further propose a regularization method named \textit{Structural Regularization} that promotes the normal convolution weights to have the desired structure that facilitates our proposed decomposition. Overall, our key contributions in this work are:
%In Sec. \ref{sec:structuredconv}, we propose \textit{Structured Convolutions} as a special case of this structure and show that it leads to a decomposition of the convolution operation into a sum-pooling operation and a significantly smaller convolution operation. We show how this decomposition can be applied to convolutional layers as well as Fully Connected / Linear layers. We further propose a regularization method named \textit{Structural Regularization} that promotes the normal convolution weights to have the desired structure which facilitates our proposed decomposition. Overall, our key contributions in this work are:
\begin{enumerate}
    \item We introduce Composite Kernel structure, which accepts an arbitrary basis in the kernel formation, leading to an efficient convolution operation. Sec.~\ref{sec:composite_definition} provides the definition.
    %\item We introduce Composite Kernel structure, which accepts an arbitrary basis in the kernel formation, leading to an efficient convolution operation. Sec. \ref{sec:composite_definition} has the definition.
    % . and we show benefits of this structure in terms of reduction in model size as well as reduction in number of multiplications required for inference. This is formally defined in Sec. \ref{sec:conv_composite}
    \item We propose Structured Convolutions, a realization of the composite kernel structure. We show that a structured convolution can be decomposed into a sum-pooling operation followed by a much smaller convolution operation. A detailed analysis is provided in Sec.~\ref{sec:decomposition}.
    %Detailed analysis of this decomposition and its application to 2D/3D kernels and fully-connected layers are given in Sec.~\ref{sec:decomposition}.
    
    %\item We propose Structured Convolutions, a realization of Composite Kernel structure. We show that a Structured Convolution can be decomposed into a sum-pooling operation followed by a significantly smaller convolution operation. Detailed analysis of this decomposition and its application to 2-D/3-D kernels and fully-connected layers is shown in Sec. \ref{sec:decomposition}.
    % \item We thoroughly investigate a special case of Composite Kernel structure and demonstrate how this structure can be applied to 2-D and 3-D kernels as well as the Linear/FC layer in a deep network. We show that imposing this structure enables a decomposition of the convolution operation into a sum-pooling operation followed by a significantly smaller convolution operation. This \textit{structural kernel decomposition} is described in Sec. \ref{sec:decomposition}.
    \item Finally, we design Structural Regularization, an effective training method to enable the structural decomposition with minimal loss of accuracy. Our process is described in Sec.~\ref{sec:regularization}.
    % Just reduced a few words for saving one line here.
    %\item Finally, we propose Structural Regularization, an effective training method to enable this decomposition with minimal loss in accuracy. This is described in Sec. \ref{sec:regularization}
\end{enumerate}

\label{section:os}

\section{Related Work}
\label{sec:related_work}
The existing literature on exploiting redundancy in deep networks can be broadly studied as follows.

\textbf{Tensor Decomposition Methods. } 
%\fp{Let's not start a paragraph with a number..} \yb{got it :)}
The work in \cite{zhang2015} proposed a Generalized SVD approach to decompose a $B \!\times\! C \!\times\! k \!\times\! k$ convolution (where $B$ and $C$ are output and input channels, and $k$ is the spatial size) into a $B' \!\times\! C \!\times\! k \!\times\! k$ convolution followed by a $B\!\times\! B' \!\times\! 1 \!\times\! 1$ convolution. Likewise, \cite{jaderberg2014speeding} introduced Spatial SVD to decompose a $k \!\times\! k$ kernel into $k \!\times\! p$ and $p \!\times\! k$ kernels. \cite{tai2015convolutional} further developed a non-iterative method for such low-rank decomposition. CP-decomposition \cite{kolda2009tensor,lebedev2014speeding} and tensor-train decomposition \cite{oseledets2011tensor,su2018tensorized,yang2017tensor} have been proposed to decompose high dimensional tensors. In our method, we too aim to decompose regular convolution into computationally lightweight units.

\textbf{Structured Pruning. }
\cite{amc,he2017,Li2016} presented channel pruning methods where redundant channels in every layer are removed. The selection process of the redundant channels is unique to every method, for instance, \cite{he2017} addressed the channel selection problem using lasso regression. Similarly, \cite{wen2016learning} used group lasso regularization to penalize and prune unimportant groups on different levels of granularity. We refer readers \cite{kuzmin2019taxonomy} for a survey of structured pruning and tensor decomposition methods. To our advantage, the proposed method in this paper does not explicitly prune, instead, our structural regularization loss imposes a form on the convolution kernels.

%\yz{\textbf{Channel Pruning.}}
%\yz{\cite{amc,he2017,Li2016} proposed channel pruning methods that remove redundant channels in every layer of a CNN. \cite{he2017} solves the channel selection problem using lasso regression. \cite{wen2016learning} uses a regularization method to learn a compact structure from a larger deep network. Our proposed structured convolution does not explicitly prune channels, but its effect, in terms of accelerating inference speed, is similar to having a channel pruning.}

\textbf{Semi-structured and Unstructured Pruning. } 
Other works \cite{lebedev2016fast,liu2018rethinking,elsen2019fast} employed block-wise sparsity (also called \textit{semi-structured} pruning) which operates on a finer level than channels. Unstructured pruning methods \cite{azarian2020learned,han2015deep,kusupati2020soft,admm} prune on the parameter-level yielding higher compression rates. However, their unstructured nature makes it difficult to deploy them on most hardware platforms.

% \cite{shrinkbench} provides a comprehensive organization and evaluation of several unstructured pruning methods.

%\yz{\textbf{Tensor Decomposition Methods.}}
%\yz{\cite{zhang2015} proposed SVD and Generalized SVD methods to decompose a $B\times C\times k\times k$ convolution kernel into a $B\times C'\times k\times k$ kernel (standard conv) and $C'\times C\times 1\times 1$ kernel (pointwise conv), with $C'<< C$. \cite{jaderberg2014speeding} introduced Spatial SVD decomposition where a $k\times k$ kernel is decomposed into $k\times p$ and $p\times k$ kernels, where $p$ is a small constant. \cite{tai2015convolutional} further developed a non-iterative method for compressing deep networks using SVD. Several other methods such as CP-decomposition \cite{kolda2009tensor,lebedev2014speeding} and tensor-train decomposition \cite{oseledets2011tensor,su2018tensorized,yang2017tensor} have been proposed to decompose high dimensional tensors. \cite{kuzmin2019taxonomy} provides a survey of structured pruning and tensor decomposition methods. Our proposed method relates to tensor decomposition in the sense that we also aim to decompose the regular convolution kernel into computationally cheaper units.

\textbf{Using Prefixed Basis. } Several works \cite{dcfnet, basisconv} applied basis representations in deep networks. Seminal works \cite{mallat2012group, sifre2013rotation} used wavelet bases as feature extractors. Choice of the basis is important, for example, \cite{dcfnet} used Fourier-Bessel basis that led to a reduction in computation complexity. In general, tensor decomposition can be seen as basis representation learning. We propose using structured binary masks as our basis, which leads to an immediate reduction in the number of multiplications.
% \cite{chan2015pcanet} proposed using PCA bases. 
% Approaches such as dictionary-learning can also be considered a form of representation learning. \cite{papyan2017convolutional} provided a theoretical analysis of deep networks by representing them as a set of convolutional sparse coding layers. 

Orthogonal to structured compression, \cite{lin2019tsm,wu2018shift,zhong2018shift} utilized shift-based operations to reduce the overall computational load. Given the high computational cost of multiplications compared to additions \cite{horowitz20141}, \cite{chen2019addernet} proposed networks where the majority of the multiplications are replaced by additions.

%\yz{\textbf{Using pre-defined basis} Recent works \cite{dcfnet, basisconv} have proposed using predefined basic masks/kernels in deep networks. The Fourier-Bessel basis used in \cite{dcfnet} lead to a significant parameter reduction. BasisConv \cite{basisconv} aims to approximate the original $D-by-D$ conv-layer by using a predefined $D-by-D$ conv-layer followed by a $1-by-by$ conv-layer. The original Seminal works like \cite{mallat2012group, sifre2013rotation} use wavelet basis in their deep networks. \cite{chan2015pcanet} proposed using PCA basis. Orthogonal to structured compression, \cite{lin2019tsm,wu2018shift,zhong2018shift} have proposed using shift-based operations to reduce the overall FLOPs count of a model. Given the high computational cost of multiplications w.r.t. additions, \cite{chen2019addernet} proposed deep networks with majority of the multiplications replaced by additions. Because these basis are hand-crafted, embedding them into a learning system could impose an unnatural bias/prior to the learning model. The overlapping sum in our structural convolution, although could be viewed as a type of pre-defined basis, is a very vanilla (unbiased) way for capturing statistics of tensor neighborhood. Overlapping sum also brings in a significant computational advantage.}

% Model Quantization \cite{bhalgat2020lsq+,krishnamoorthi2018quantizing,dfq} methods also reduce redundancy by representing weights in lower bit-widths. But quantization can also be applied to activations to further improve run-time efficiency.

\section{Composite Kernels}
\label{sec:composite_definition}

We first give a definition that encompasses a wide range of structures for convolution kernels.
\begin{definition}
For $\mathcal{R}^{C \!\times\! N \!\times\! N}$, a Composite Basis $\mathcal{B}=\{\beta_1, \beta_2,...,\beta_M\}$ is a linearly independent set of binary tensors of dimension $C \!\times\! N \!\times\! N$ as its basis elements. That is, $\beta_m \in \mathcal{R}^{C \!\times\! N \!\times\! N}$, each element $\beta_{mijk}$ of $\beta_m\in\{0,1\}$, and $\sum_{m=1}^M \alpha_m\beta_m = 0 \;\; \text{iff} \;\; \alpha_m=0 \; \forall m$.
% % \vspace*{-1mm}
% \begin{gather*}
%     \beta_m \in \mathcal{R}^{C\times N\times N}, \quad \beta_{mijk} \in \{0,1\} \; \forall \; i \in \{1,..,C\}, j,k \in \{1,..,N\} \quad \text{and} \\ \sum_{m=1}^M \alpha_m\beta_m = 0 \iff \forall \; m, \alpha_m=0
% \end{gather*}
\end{definition}
The linear independence condition implies that $M\leq CN^2$. Hence, the basis spans a subspace of $\mathcal{R}^{C \!\times\! N \!\times\! N}$. The speciality of the Composite Basis is that the basis elements are binary, which leads to an immediate reduction in the number of multiplications involved in the convolution operation.
%The structure of a Composite Kernel is defined by the Composite Basis.\\

\begin{definition}
A kernel $W\in \mathcal{R}^{C \!\times\! N \!\times\! N}$ is a Composite Kernel if it is in the subspace of the Composite Basis. That is, it can be constructed as a linear combination of the elements of $\mathcal{B}$: $\exists$ $\boldsymbol\alpha=[\alpha_1,..,\alpha_M]$ such that $W \coloneqq \sum_{m=1}^M \alpha_m \beta_m$.
%A kernel $W\in \mathcal{R}^{C\times N\times N}$ is a Composite Kernel if it can be constructed as a linear combination of a Composite Basis, i.e. $W\coloneqq \sum_{m=1}^M \alpha_m \beta_m$ for some $\boldsymbol\alpha=[\alpha_1,..,\alpha_M]$. 
\end{definition}

Note that, the binary structure of the underlying Composite Basis elements defines the structure of the Composite Kernel. Fig.~\ref{fig:OS_illustration} shows a $3 \!\times\! 3$ Composite Kernel (with $C=1, N=3$) constructed using different examples of a Composite Basis. In general, the underlying basis elements could have a more random structure than what is demonstrated in those examples shown in Fig.~\ref{fig:OS_illustration}. 

%For illustration, we show the same Composite Kernel in Fig.~\ref{fig:OS_illustration}(a) as a linear combination of the Composite Basis elements:
%\begin{figure}[h]
%    \centering
%    \includegraphics[width=0.75\textwidth]{Figures/OS_definition.png}
%\end{figure}

Conventional kernels (with no restrictions on DOF) are just special cases of Composite Kernels, where $M=CN^2$ and each basis element has only one nonzero element in its $C \!\times\! N \!\times\! N$ grid.

\subsection{Convolution with Composite Kernels}
\label{sec:conv_composite}
Consider a convolution with a Composite Kernel of size $C \!\times\! N \!\times\! N$, where $N$ is the spatial size and $C$ is the number of input channels. To compute an output, this kernel is convolved with a $C \!\times\! N \!\times\! N$ volume of the input feature map. Let's call this volume $X$. Therefore, the output at this point will be:
\begin{equation}
    X * W = X * \sum_{m=1}^M\alpha_m\beta_m = \sum_{m=1}^M\alpha_m(X*\beta_m) = \sum_{m=1}^M\alpha_m\, sum(X\bullet \beta_m) = \sum_{m=1}^M\alpha_m E_m
    \label{eq:conv_OS}
\end{equation}

% With a Composite Kernel, the convolution operations can be performed with much lesser number of multiplications and additions. One way of performing a convolution is to stride the convolution kernel across the input feature map and calculate the convolution at every point. Here and throughout the paper, we will assume that the kernel is already flipped. So, the convolution at a particular stride effectively becomes a inner product operation.

% Consider a convolution with a Composite Kernel $W$, i.e. $W=\sum_{m=1}^M\alpha_m\beta_m$ for some underlying basis $\mathcal{B}$. For a particular stride of the convolution operation, let the input patch be $X$. Hence, the output $Y$ at this point can be computed as:

where `$*$' denotes convolution, `$\bullet$' denotes element-wise multiplication. Since $\beta_m$ is a binary tensor, $sum(X\bullet \beta_m)$ is same as adding the elements of $X$ wherever $\beta_m=1$, thus no multiplications are needed. Ordinarily, the convolution $X*W$ would involve $CN^2$ multiplications and $CN^2-1$ additions. In our method, we can trade multiplications with additions. From (\ref{eq:conv_OS}), we can see that we only need $M$ multiplications and the total number of additions becomes:
\begin{align}
\text{Num additions} = \sum_{m=1}^M \underbrace{(sum(\beta_m)-1)}_{\text{from } sum(X\bullet \beta_m)} \;\; + \underbrace{(M-1)}_{\text{from } \sum\alpha_m E_m} = \sum_{m=1}^M sum(\beta_m) -1
\end{align}

Depending on the structure, number of the additions \textit{per output} can be larger than $CN^2-1$. For example, in Fig.~\ref{fig:OS_illustration}(b) where $C=1, N=3, M=4$, we have $\sum_m sum(\beta_m)-1=15 > CN^2-1=8$). In Sec. \ref{sec:num_ops}, we show that the number of additions can be amortized to as low as $M-1$.

\section{Structured Convolutions}
\label{sec:structuredconv}
% In this section, we define Structured Kernels as a realization of Composite Kernels. , will be investigated further.

\begin{figure}[t]
\centering
\begin{minipage}{.53\textwidth}
  \centering
  \includegraphics[width=.85\linewidth]{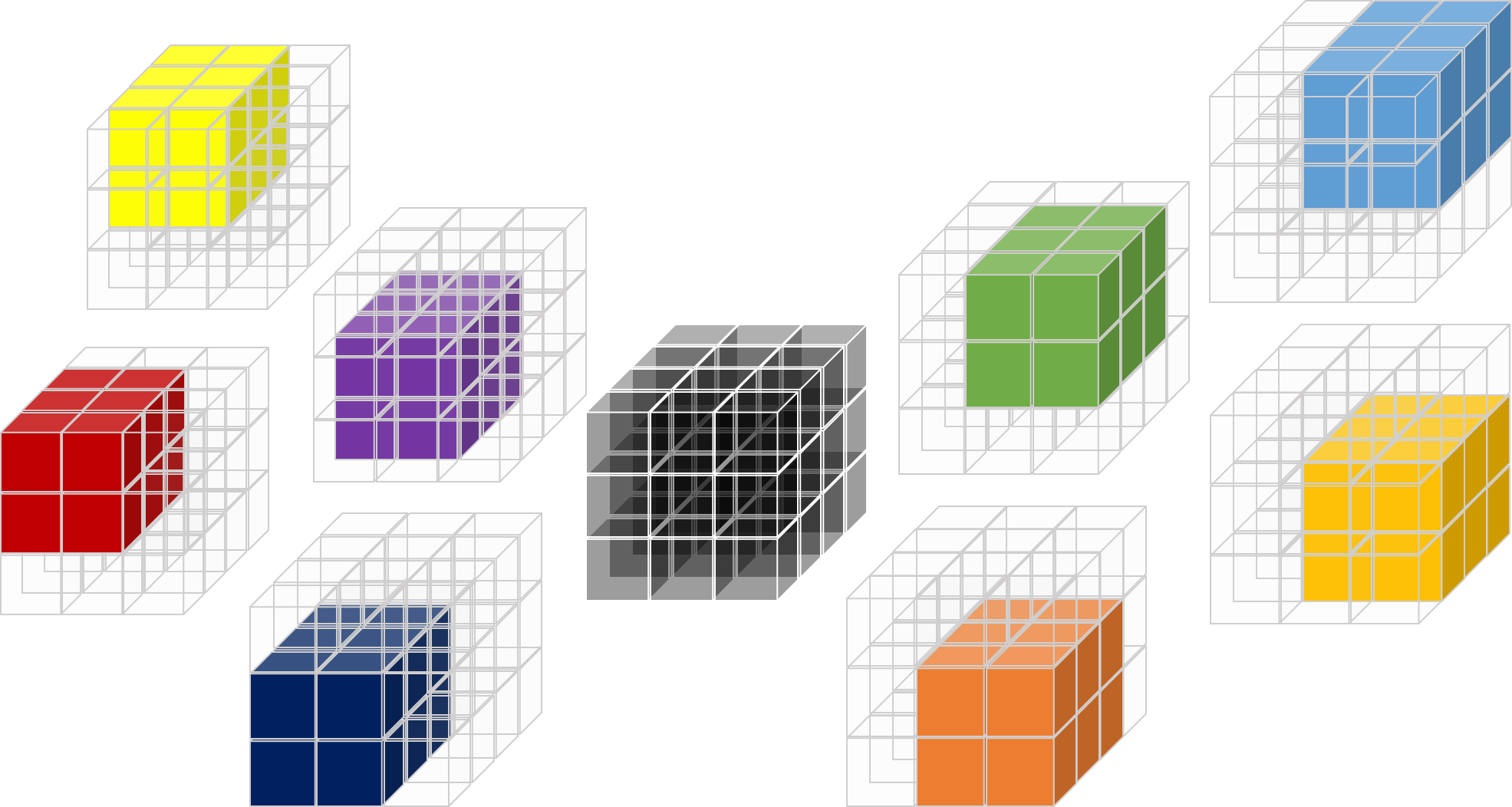}
  \captionof{figure}{\small{$4 \!\times\! 3\times 3$ structured kernel constructed with 8 basis elements each having a $3 \!\times\! 2 \!\times\! 2$ patch of $1$'s. \\ \textit{Figure best viewed in color.}}}
  \label{fig:structured_kernel}
\end{minipage} \hfill
\begin{minipage}{.43\textwidth}
  \centering
  \includegraphics[width=.9\linewidth]{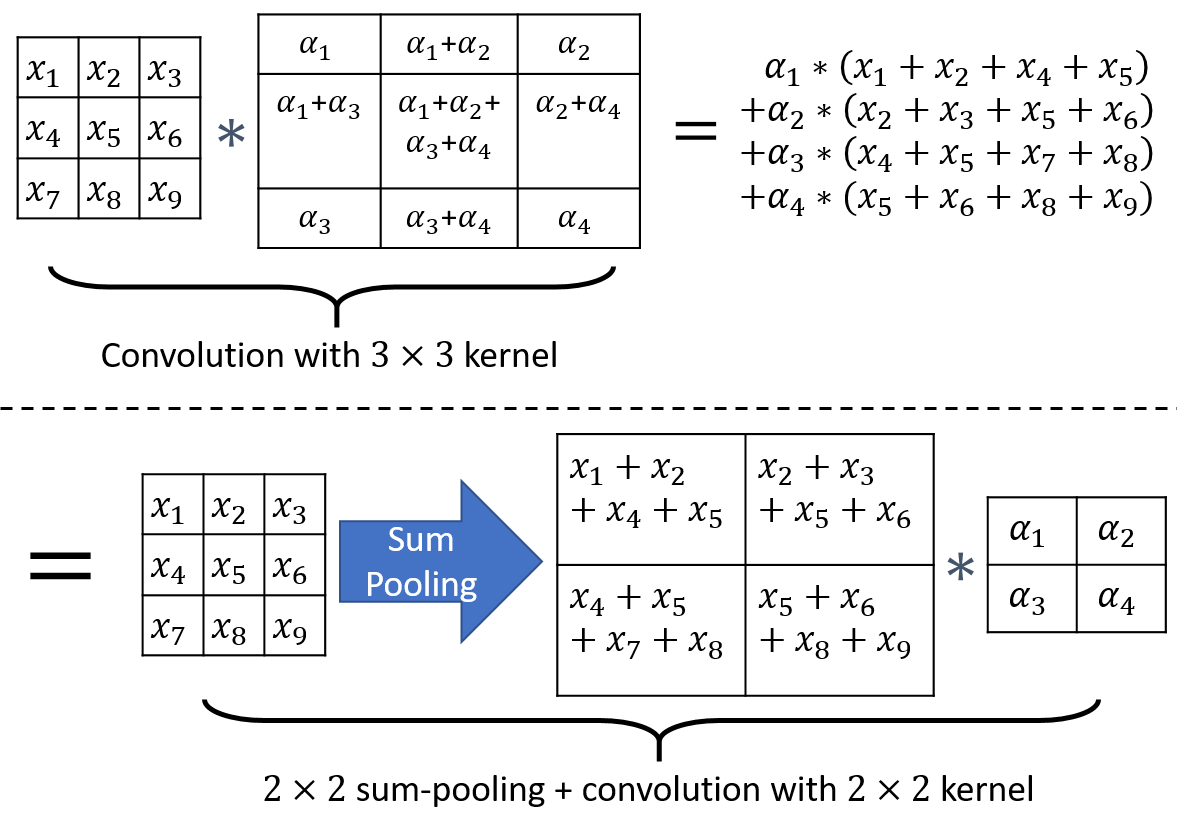}
  \captionof{figure}{\small{$3 \!\times\! 3$ structured convolution is equivalent to $2 \!\times\! 2$ sum-pooling + $2 \!\times\! 2$ convolution.}}
  \label{fig:conv_3x3}
\end{minipage}
\end{figure}

% \begin{figure}[t]
%     \centering
%     \begin{minipage}{0.5\textwidth}
%         \centering
%         \includegraphics[width=0.98\textwidth]{Figures/3D_structured.png}
%         (a) \small{$4\times 3\times 3$ structured kernel constructed with 8 basis kernels each having a $3\times 2 \times 2$ patch of $1$'s. \\ \textit{Figure best viewed in color.}} \\
%         % $C, N, D, k = 4, 3, 2, 2 \Rightarrow C-D+1=3, N-k+1=2$
%         % first figure itself
%     \end{minipage}\hfill
%     \begin{minipage}{0.45\textwidth}
%         \centering
%         \includegraphics[width=0.98\textwidth]{Figures/conv_3x3_alt.png}
%         (b) \small{Decomposition of convolution with a $3\times 3$ structured kernel}\\
%     \end{minipage}
%     \caption{}
%     \label{fig:structured_kernel}
% \end{figure}

% \subsection{Definition of Structured Kernels}
\begin{definition}
A kernel in $\mathcal{R}^{C \!\times\! N \!\times\! N}$ is a Structured Kernel if it is a Composite Kernel with $M=cn^2$ for some $1\leq n \leq N, 1\leq c \leq C$, and if each basis tensor $\beta_m$ is made of a $(C-c+1) \times (N-n+1)\times (N-n+1)$ cuboid of $1$'s, while rest of its coefficients being $0$. 
\end{definition}

A Structured Kernel is characterized by its dimensions $C, N$ and its underlying parameters $c, n$. Convolutions performed using Structured Kernels are called Structured Convolutions. 

Fig.~\ref{fig:OS_illustration}(b) depicts a 2D case of a $3 \!\times\! 3$ structured kernel where $C=1, N=3, c=1, n=2$. As shown, there are $M=cn^2=4$ basis elements and each element has a $2 \!\times\! 2$ sized patch of $1$'s. 

Fig.~\ref{fig:structured_kernel} shows a 3D case where $C=4, N=3, c=2, n=2$. Here, there are $M=cn^2=8$ basis elements and each element has a $3 \!\times\! 2 \!\times\! 2$ cuboid of $1$'s. Note how these cuboids of $1$'s (shown in colors) cover the entire $4 \!\times\! 3 \!\times\! 3$ grid.

\subsection{Decomposition of Structured Convolutions}
\label{sec:decomposition}
% A structured convolution can be decomposed into a sum-pooling operation and a smaller convolution operation. To understand this, 

% For structured kernels, each $\beta_m$ has a $(C-D+1)\times(N-k+1)\times(N-k+1)$ cuboid of $1$'s in a unique position in the $C\times N\times N$ grid. Hence, $X*\beta_m$ in Eq. (\ref{eq:conv_OS}) corresponds to a unique stride of a sum-pooling operation with kernel size of $(C-D+1)\times(N-k+1)\times(N-k+1)$.

A major advantage of defining Structured Kernels this way is that all the basis elements are just shifted versions of each other (see Fig.~\ref{fig:structured_kernel} and Fig.~\ref{fig:OS_illustration}(b)). 
This means, in Eq.~(\ref{eq:conv_OS}), if we consider the convolution $X*W$ for the \textit{entire} feature map $X$, the summed outputs $X*\beta_m$ for all $\beta_m$'s are actually the same (except on the edges of $X$). As a result, the outputs $\{X*\beta_1,...,X*\beta_{cn^2}\}$ can be computed using a \textit{single sum-pooling operation} on $X$ with a kernel size of $(C-c+1)\times(N-n+1)\times(N-n+1)$.
Fig.~\ref{fig:conv_3x3} shows a simple example of how a convolution with a $3 \!\times\! 3$ structured kernel can be broken into a $2 \!\times\! 2$ sum-pooling followed by a $2 \!\times\! 2$ convolution with a kernel made of $\alpha_i$'s.

Furthermore, consider a convolutional layer of size $C_{out}\!\times\! C \!\times\! N \!\times\! N$ that has $C_{out}$ kernels of size $C\smalltimes N\smalltimes N$. In our design, the same underlying basis $\mathcal{B}=\{\beta_1,...,\beta_{cn^2}\}$ is used for the construction of all $C_{out}$ kernels in the layer. Suppose any two structured kernels in this layer with coefficients $\boldsymbol\alpha^{(1)}$ and $\boldsymbol\alpha^{(2)}$, i.e. $W_1=\sum_m\alpha_m^{(1)}\beta_m$ and $W_2=\sum_m\alpha_m^{(2)}\beta_m$. The convolution output with these two kernels is respectively, $X*W_1=\sum_m\alpha_m^{(1)}(X*\beta_m)$ and $X*W_2=\sum_m\alpha_m^{(2)}(X*\beta_m)$. We can see that the computation $X*\beta_m$ is common to all the kernels of this layer. Hence, the sum-pooling operation only needs to be computed once and then reused across all the $C_{out}$ kernels.

\begin{figure}[t]
    \centering
    \includegraphics[width=\textwidth]{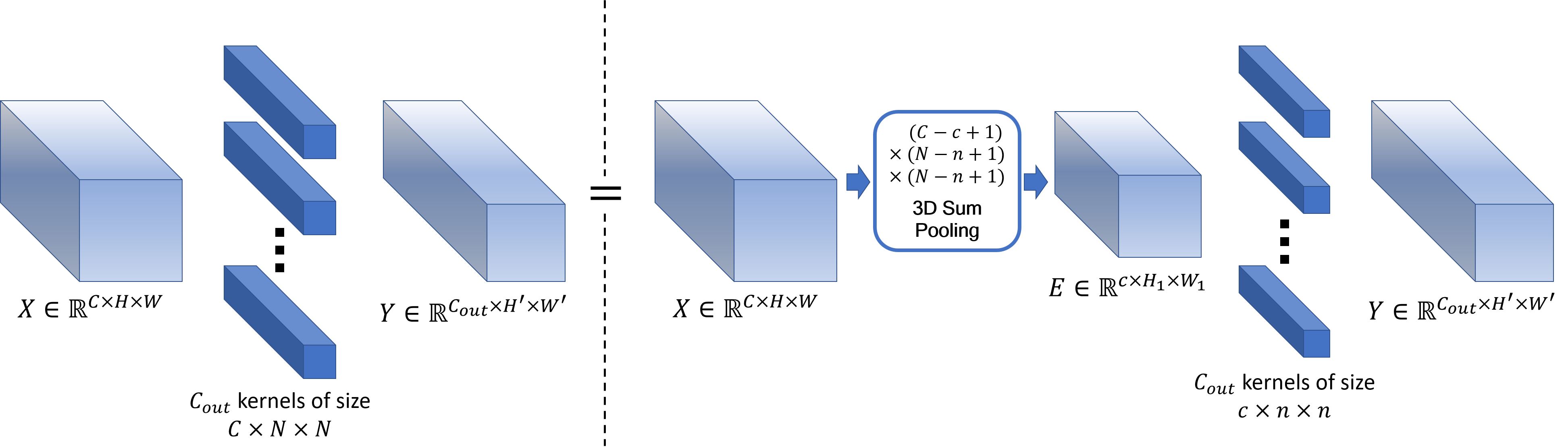}
    \caption{Decomposition of Structured Convolution. On the left, the conventional operation of a structured convolutional layer of size $C_{out} \smalltimes C \smalltimes N \smalltimes N$ is shown. On the right, we show that it is equivalent to performing a 3D sum-pooling followed by a convolutional layer of size $C_{out} \smalltimes c \smalltimes n \smalltimes n$.}
    \label{fig:structural_decomposition}
\end{figure}

A Structured Convolution can thus be decomposed into a sum-pooling operation and a smaller convolution operation with a kernel composed of $\alpha_i$'s. Fig.~\ref{fig:structural_decomposition} shows the decomposition of a general structured convolution layer of size $C_{out} \!\times\! C \!\times\! N \!\times\! N$.

Notably, standard convolution ($C\smalltimes N\smalltimes N$), depthwise convolution ($1 \smalltimes N \smalltimes N$), and pointwise convolution ($C \smalltimes 1 \smalltimes 1$) kernels can all be constructed as 3D structured kernels, which means that this decomposition can be widely applied to existing architectures. See supplementary material for more details on applying the decomposition to convolutions with arbitrary \textit{stride, padding, dilation}.

% Until now, we have considered only a single stride of the convolution operation $X*W$. But the above breakdown holds even when we consider the entire convolution operation, i.e. \textit{all strides together} and \textit{all $C_{out}$ kernels of a Conv layer considered together}.\yb{Need to add some more details here.} Hence, for a general $C\times N\times N$ structured convolution with underlying parameters $\{D, k\}$ and $C_{out}$ output channels, the following decomposition holds:

% \begin{figure}[h]
%     \centering
%     \includegraphics[width=0.98\textwidth]{Figures/structural_decomposition.png}
%     \caption{3D structural decomposition of a structured convolution.}
%     \label{fig:structural_decomposition}
% \end{figure}

% Since, 2D structured kernel is a special case of 3D structured kernel where $C=D=1$, Fig. \ref{fig:structural_decomposition_2D} shows how the 2D structural decomposition would be implemented:
% \begin{figure}[h]
%     \centering
%     \includegraphics[width=0.98\textwidth]{Figures/structural_decomposition_2D.png}
%     \caption{2D structural decomposition of a structured convolution. This is a special case of Fig. \ref{fig:structural_decomposition} when $C=D=1$.}
%     \label{fig:structural_decomposition_2D}
% \end{figure}

\subsection{Reduction in Number of Parameters and Multiplications/Additions}
\label{sec:num_ops}

The sum-pooling component after decomposition requires no parameters. Thus, the total number of parameters in a convolution layer get reduced from $C_{out}CN^2$ (before decomposition) to $C_{out}cn^2$ (after decomposition). The sum-pooling component is also free of multiplications. Hence, only the smaller $c \smalltimes n \smalltimes n$ convolution contributes to multiplications after decomposition. %Exact sizes of the input and output tensors in Fig.~\ref{fig:structural_decomposition} can help calculating number of operations before and after decomposition. 

Before decomposition, computing every output element in feature map $Y \in \mathcal{R}^{C_{out} \smalltimes H' \smalltimes W'}$ involves $CN^2$ multiplications and $CN^2-1$ additions. Hence, total multiplications involved are $CN^2 C_{out}H'W'$ and total additions involved are $(C N^2-1) C_{out}H'W'$.

After decomposition, computing every output element in feature map $Y$ involves $cn^2$ multiplications and $cn^2-1$ additions. Hence, total multiplications and additions involved in computing $Y$ are $cn^2 C_{out}H'W'$ and $(cn^2-1) C_{out}H'W'$ respectively. Now, computing every element of the intermediate sum-pooled output $E \in \mathcal{R}^{c \smalltimes H_1 \smalltimes W_1}$ involves $((C-c+1)(N-n+1)^2-1)$ additions. Hence, the overall total additions involved can be written as: \[C_{out} \left( \frac{((C-c+1)(N-n+1)^2-1) cH_1W_1}{C_{out}} + (c n^2-1) H'W'\right)\]

% \begin{table}[h]
%     \small
%     \centering
%     \begin{tabular}{ccc}
%         \toprule
%         {} & \shortstack{Conventional convolution\\(before decomposition)} & \shortstack{Sum-pooling + Smaller convolution\\(after decomposition)} \\
%         \midrule
%         \# Parameters & $C_{out} C N^2$  & $C_{out} c n^2$ \\
%         \# Multiplications & $C N^2 \times C_{out}H'W'$ & $c n^2 \times C_{out}H'W'$ \\
%         \# Additions & $(C N^2-1)\times C_{out}H'W'$ & \makecell{$((C-D+1)(N-k+1)^2-1)\times cH_1W_1$\\$+ (c n^2-1) \times C_{out}H'W'$} \\
%         \bottomrule
%     \end{tabular}
% \end{table}

% \fp{these table is so not clear that readers would not appreciate how much reduction we achieve.... LEt's think how to simplify (you can move some stuff to supplementary) and make the compute gain clearer} \yb{I agree it can be explained better. But Saurabh gave me feedback that this section was crystal clear to him.. Let me try make it better}

We can see that the number of parameters and number of multiplications have both reduced by a factor of $\mathbf{{cn^2}/{CN^2}}$. And in the expression above, if $C_{out}$ is large enough, the first term inside the parentheses gets amortized and the number of additions $\approx (c n^2-1) C_{out}H'W'$. As a result, the number of additions also reduce by approximately the same proportion $\approx cn^2/CN^2$. We will refer to $CN^2/cn^2$ as the compression ratio from now on. 

Due to amortization, the additions \textit{per output} are $\approx cn^2-1$, which is basically $M-1$ since $M=cn^2$. %This can be contrasted with the additions calculated in Sec.~\ref{sec:conv_composite} for general Composite Kernels.

\subsection{Extension to Fully Connected layers} 
%\jl{ If space is permitted, I encourage to use an illustration to demonstrate this important contribution for 1x1/FC. }

For image classification networks, the last fully connected layer (sometimes called linear layer) dominates w.r.t. the number of parameters, especially if the number of classes is high. The structural decomposition can be easily extended to the linear layers by noting that a matrix multiplication is the same as performing a number of $1 \smalltimes 1$ convolutions on the input. Consider a kernel $W \in \mathcal{R}^{P\times Q}$ and input vector $X \in \mathcal{R}^{Q \times 1}$. The linear operation $WX$ is mathematically equivalent to the $1\times 1$ convolution $unsqueezed(X)*unsqueezed(W)$, where $unsqueezed(X)$ is the same as $X$ but with dimensions $Q\times 1\times 1$ and $unsqueezed(W)$ is the same as $W$ but with dimensions $P\times Q\times 1\times 1$. In other words, each row of $W$ can be considered a $1\times 1$ convolution kernel of size $Q\times 1\times 1$.

\begin{figure}[h]
    \centering
    \includegraphics[width=0.98\textwidth]{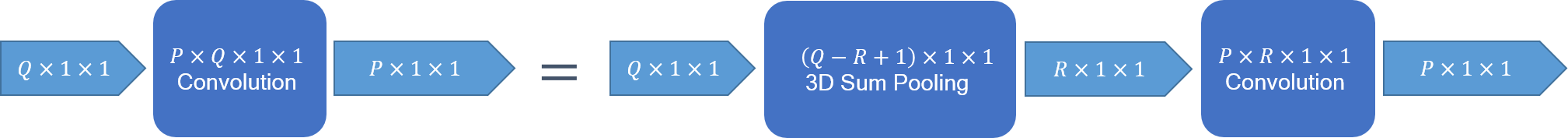}
    \caption{Structural decomposition of a matrix multiplication.}
    \label{fig:structural_decomposition_matrix}
\end{figure}

Now, if each of these kernels (of size $Q\times 1\times 1$) is \textit{structured} with underlying parameter $R$ (where $R\leq Q$), then the matrix multiplication operation can be structurally decomposed as shown in Fig.~\ref{fig:structural_decomposition_matrix}.

Same as before, we get a reduction in both the number of parameters and the number of multiplications by a factor of $\mathbf{R/Q}$, as well as the number of additions by a factor of $\frac{R(Q-R)+(PR-1)P}{(PQ-1)P}$.

\section{Imposing Structure on Convolution Kernels}
To apply the structural decomposition, we need the weight tensors to be structured. In this section, we propose a method to impose the desired structure on the convolution kernels via training.

From the definition, $W=\sum_{m}\alpha_m\beta_m$, we can simply define matrix $A\in \mathcal{R}^{CN^2 \smalltimes cn^2}$ such that its $i^{th}$ column is the vectorized form of $\beta_i$. Hence, $\text{vectorized}(W)=A \cdot\boldsymbol\alpha$, where $\boldsymbol\alpha=[\alpha_1, ...,\alpha_{cn^2}]$. 
% This is to say that structured kernels are characterized by a structure matrix $A$.

Another way to see this is from structural decomposition. We may note that the $(C-c+1) \smalltimes (N-n+1) \smalltimes (N-n+1)$ sum-pooling can also be seen as a convolution with a kernel of all $1$'s; we refer to this kernel as $\mathbf{1}_{(C-c+1) \smalltimes (N-n+1) \smalltimes (N-n+1)}$. Hence, the structural decomposition is: 
\[X*W = X*\mathbf{1}_{(C-c+1) \smalltimes (N-n+1) \smalltimes (N-n+1)}* \boldsymbol\alpha_{c \smalltimes n \smalltimes n}\] 

That implies, $W=\mathbf{1}_{(C-c+1) \smalltimes (N-n+1) \smalltimes (N-n+1)}* \boldsymbol\alpha_{c \smalltimes n \smalltimes n}$. Since the stride of the sum-pooling involved is $1$, this can be written in terms of a matrix multiplication with a Topelitz matrix \cite{strang1986proposal}: 
\[\text{vectorized}(W) = \text{Topelitz}(\mathbf{1}_{(C-c+1) \smalltimes (N-n+1) \smalltimes (N-n+1)}) \cdot \text{vectorized}(\boldsymbol\alpha_{c \smalltimes n \smalltimes n})\] 

Hence, the structure matrix $A$ referred above is basically $\text{Topelitz}(\mathbf{1}_{(C-c+1) \smalltimes (N-n+1) \smalltimes (N-n+1)})$.

% A simple pseudocode to generate this $A$ matrix has been added to the Supplementary Material.

\subsection{Training with Structural Regularization}
\label{sec:regularization}
Now, for a structured kernel $W$ characterized by $\{C, N, c, n\}$, there exists a $cn^2$ length $\boldsymbol\alpha$ such that $W=A\boldsymbol\alpha$. Hence, a structured kernel $W$ satisfies the property: $W=AA^+W$, where $A^+$ is the Moore-Penrose inverse \cite{ben2003generalized} of $A$. Based on this, we propose training a deep network with a Structural Regularization loss that can gradually push the deep network's kernels to be structured \textit{via training}:
\begin{equation}
    \mathcal{L}_{total} = \mathcal{L}_{task} + \lambda\sum_{l=1}^L\frac{\norm{(I-A_lA_l^+)W_l}_F}{\norm{W_l}_F}
    \label{eq:structural_regularization_loss}
\end{equation}
where $\norm{\cdot}_F$ denotes Frobenius norm and $l$ is the layer index. To ensure that regularization is applied uniformly to all layers, we use $\norm{W}_F$ normalization in the denominator. It also stabilizes the performance of the decomposition w.r.t $\lambda$. The overall proposed training recipe is as follows:

\textbf{Proposed Training Scheme:} 
% \vspace{-5pt}
\begin{itemize}[topsep=0pt]
    \item \emph{Step 1}: Train the original architecture with the Structural Regularization loss.
    \begin{itemize}[topsep=0pt]
        \item After Step 1, all weight tensors in the deep network will be \textit{almost} structured.
    \end{itemize}
    \item \emph{Step 2}: Apply the decomposition on every layer and compute $\alpha_l=A_l^+W_l$.
    \begin{itemize}[topsep=0pt]
        \item This results in a smaller and more efficient decomposed architecture with $\alpha_l$'s as the weights. Note that, every convolution / linear layer from the original architecture is now replaced with a sum-pooling layer and a smaller convolution / linear layer.
    \end{itemize}
\end{itemize}

% Fig. \ref{fig:training_recipe}. 

% If $(I-AA^+)W=0$ for all kernels, then we say that the decomposition $\alpha=A^+W$ is "exact", meaning that the decomposed architecture (with $\alpha$'s as weights) is mathematically equivalent to the original architecture before decomposition.
% \begin{figure}[t]
%     \centering
%     \includegraphics[width=0.98\textwidth]{Figures/training_recipe.png}
%     \caption{Overall training recipe for Structural Decomposition. \yb{Will remove this figure, not necessary.}}
%     \label{fig:training_recipe}
% \end{figure}

% \subsection{Significance of $\lambda$}

The proposed scheme trains the architecture with the original $C \smalltimes N \smalltimes N$ kernels in place but with a structural regularization loss. The structural regularization loss imposes a restrictive $cn^2$ degrees of freedom while training but in a \textit{soft} or gradual manner (depending on $\lambda$):
\begin{enumerate}[leftmargin=*,topsep=0pt,itemsep=0pt]
    \item If $\lambda=0$, it is the same as normal training with no structure imposed.
    \item If $\lambda$ is very high, the regularization loss will be heavily minimized in early training iterations. Thus, the weights will be optimized in a restricted $cn^2$ dimensional subspace of $\mathcal{R}^{C \smalltimes N \smalltimes N}$.
    \item Choosing a moderate $\lambda$ gives the best tradeoff between structure and model performance.
\end{enumerate}

We talk about training implementation details for reproduction, such as hyperparameters and training schedules, in Supplementary material, where we also show our method is robust to the choice of $\lambda$.

% An \textbf{alternative training scheme} could be to first train the original architecture without structural regularization, i.e., normal training. Then, decompose the pretrained weights using $\alpha = A^+W$ and finetune the decomposed architecture.
% Since the architecture is already decomposed during finetuning,
% each weight tensor is optimized in a restricted $cn^2$ dimensional subspace of $\mathcal{R}^{C \smalltimes N \smalltimes N}$.

% Our proposed training scheme is used to generate all the results in Sec.~\ref{results}. We also provide additional results in Supplementary material that are obtained from the alternative training scheme.

\section{Experiments}
\label{results}

We apply structured convolutions to a wide range of architectures and analyze the performance and complexity of the decomposed architectures. We evaluate our method on ImageNet \cite{ILSVRC15} and CIFAR-10 \cite{cifar} benchmarks for image classification and Cityscapes \cite{cordts2016cityscapes} for semantic segmentation.  
%We apply the proposed method to a wide range of CNN architectures and investigate the performance as well as complexity of the decomposed architectures. We empirically evaluate our method on ImageNet \cite{ILSVRC15} and CIFAR-10 \cite{cifar} datasets for image classification and the Cityscapes \cite{cordts2016cityscapes} dataset for semantic segmentation. Using different configurations of $\{c,n\}$ per layer, we get \textit{structured} versions of these architectures with varying levels of reduction in the model size and multiply/add count. 

\begingroup
\setlength{\tabcolsep}{2pt}
\begin{table}[h]
    \begin{minipage}{.5\linewidth}
        \begin{minipage}{\linewidth}
        \small
        \centering
        \caption{Results: ResNets on CIFAR-10}
        \begin{tabular}{lcccc}
            \toprule
            \shortstack{Architecture} & \shortstack{Adds\\($\times 10^6$)} & \shortstack{Mults\\($\times 10^6$)} & \shortstack{Params\\($\times 10^6$)} & \shortstack{Acc. \\ (in \%)}\\
            \midrule
            \midrule
            ResNet56 & $125.49$ & $126.02$ & $0.85$ & ${93.03}$  \\
            \textbf{Struct-56-A (ours)} & ${63.04}$ & ${62.10}$ & ${0.40}$ & $92.65$  \\
            \textbf{Struct-56-B (ours)} & $48.49$ & $33.87$ & $0.21$ & $91.78$ \\
            % \vspace*{1mm}
            % \hdashline
            \arrayrulecolor{black!30}\midrule
            Ghost-Res56 \cite{ghostnet} & $63.00$ & $63.00$ & $0.43$ & $92.70$ \\
            ShiftRes56-6 \cite{wu2018shift} & ${51.00}$ & ${51.00}$ & $0.58$ & $92.69$\\
            AMC-Res56 \cite{amc} & $63.00$ & $63.00$ & -- & $91.90$\\
            % ShiftResNet56-3 & $27.00$ & $27.00$ & $0.29$ & $92.11$\\

            \arrayrulecolor{black}\midrule
            \midrule
            ResNet32 & $68.86$ & $69.16$ & $0.46$ &  ${92.49}$ \\
            \textbf{Struct-32-A (ours)} & ${35.59}$ & ${35.09}$ & ${0.22}$ & $92.07$  \\
            \textbf{Struct-32-B (ours)} & $26.29$ & $18.18$ & $0.11$ & $90.24$  \\
            \midrule
            \midrule
            ResNet20 & $40.55$ & $40.74$ & $0.27$ & ${91.25}$ \\
            %\arrayrulecolor{black!30}\midrule
            % ShiftResNet20-3 & $13.00$ & $13.00$ & $0.10$ & $90.08$\\
            \textbf{Struct-20-A (ours)} & ${20.77}$ & ${20.42}$ & ${0.13}$ & $91.04$ \\
            \textbf{Struct-20-B (ours)} & ${15.31}$ & ${10.59}$ & $0.067$ & $88.47$  \\
            % \hdashline
            \arrayrulecolor{black!30}\midrule
            ShiftRes20-6 \cite{wu2018shift} & $23.00$ & $23.00$ & $0.19$ & $90.59$\\
            
            \arrayrulecolor{black}\bottomrule 
        \end{tabular}
        \label{table:cifar_resnet}
        \end{minipage}
        % \vspace*{10pt}
        % \bigskip
        \begin{minipage}{\linewidth}
        \small
        \centering
        \caption{Results: MobileNetV2 on ImageNet}
        \begin{tabular}{lcccc}
            \toprule
            \shortstack{Architecture} & \shortstack{Adds\\($\times 10^6$)} & \shortstack{Mults\\($\times 10^6$)} & \shortstack{\footnotesize{Params}\\($\times 10^6$)} & \shortstack{Acc. \\ (in \%)}\\
            \midrule
            \midrule
            MobileNetV2 & $0.30$ & $0.31$ & $3.50$ & ${72.19}$\\
            \textbf{Struct-V2-A (ours)} & ${0.26}$ & ${0.23}$ & ${2.62}$ & $71.29$ \\
            \textbf{Struct-V2-B (ours)} & $0.29$ & $0.17$ & ${1.77}$ & ${64.93}$ \\
            % \hdashline
            \arrayrulecolor{black!30}\midrule
            AMC-MV2 \cite{amc} & $0.21$ & $0.21$ & -- & $70.80$ \\
            ChPrune-MV2-1.3x & $0.22$ & ${0.23}$ & $2.58$ & $68.99$ \\
            Slim-MV2 \cite{slimmable} & $0.21$ & $0.21$ & $2.60$ & $68.90$ \\
            WeightSVD 1.3x & ${0.22}$ & ${0.23}$ & ${2.45}$ & $67.48$ \\
            %\midrule
            ChPrune-MV2-2x & ${0.15}$ & ${0.15}$ & $1.99$ & $63.82$ \\
            % WeightSVD 2x & ${0.15}$ & ${0.15}$ & ${1.44}$ & $60.11$ \\
            
            \arrayrulecolor{black}\bottomrule 
        \end{tabular}
        \label{table:imagenet_mv2}
        \end{minipage}
    \end{minipage}
    \hfill
    \begin{minipage}{.5\linewidth}
    \begin{minipage}{\linewidth}
    \small
    \centering
    \caption{Results: ResNets on ImageNet}
    \begin{tabular}{lcccc}
        \toprule
        \shortstack{Architecture} & \shortstack{Adds\\($\times 10^9$)} & \shortstack{Mults\\($\times 10^9$)} & \shortstack{Params\\($\times 10^6$)} & \shortstack{Acc. \\ (in \%)} \\
        \midrule
        \midrule
        ResNet50 & $4.09$ & $4.10$ & $25.56$ & ${76.15}$  \\
        \textbf{Struct-50-A (ours)} & ${2.69}$ & ${2.19}$ & ${13.49}$ & $75.65$  \\
        \textbf{Struct-50-B (ours)} & $1.92$ & $1.38$ & ${8.57}$ & $73.41$  \\
        \arrayrulecolor{black!30}\midrule
        % Versatile-R50 \cite{versatile} & $3.20$ & $3.20$ & $19.00$ & $75.50$\\
        ChPrune-R50-2x \cite{he2017} & ${2.04}$ & ${2.05}$ & $17.89$ & $75.44$ \\
        WeightSVD-R50 \cite{zhang2016accelerating} & $2.04$ & $2.05$ & $13.40$ & $75.12$ \\
        Ghost-R50 ({s=2}) \cite{ghostnet} & $2.20$ & $2.20$ & $13.00$ & $75.00$ \\
        % Slim-R50 $0.75$x \cite{slimmable} & $2.30$ & $2.30$ & $14.70$ & $74.90$\\
        Versatile-v2-R50 \cite{versatile} & $3.00$ & $3.00$ & $11.00$ & $74.50$\\
        ShiftResNet50 \cite{wu2018shift} & -- & -- & $11.00$ & $73.70$ \\
        Slim-R50 $0.5$x \cite{slimmable} & $1.10$ & $1.10$ & $6.90$ & $72.10$\\
        
        % ChPrune-R50-4x & ${1.02}$ & ${1.02}$ & $10.04$ & $68.95$ \\
        % WeightSVD-R50-4x & ${1.02}$ & ${1.03}$ & $6.10$ & $71.55$ \\

        \arrayrulecolor{black}\midrule
        \midrule
        ResNet34 & $3.66$ & $3.67$ & $21.80$ &  ${73.30}$ \\
        \textbf{Struct-34-A (ours)} & ${1.71}$ & ${1.71}$ & ${9.82}$ & $72.81$  \\
        \textbf{Struct-34-B (ours)} & ${1.47}$ & ${1.11}$ & ${5.60}$ & $69.44$  \\
        \arrayrulecolor{black}\midrule
        \midrule
        ResNet18 & $1.81$ & $1.82$ & $11.69$ & ${69.76}$ \\
        \textbf{Struct-18-A (ours)} & $0.88$ & $0.89$ & $5.59$ & $69.13$ \\
        \textbf{Struct-18-B (ours)} & $0.79$ & ${0.63}$ & ${3.19}$ & ${66.19}$  \\
        \arrayrulecolor{black!30}\midrule
        WeightSVD-R18 \cite{zhang2016accelerating}  & $0.89$ & $0.90$ & $5.83$ & $67.71$ \\
        ChPrune-R18-2x \cite{he2017} & $0.90$ & $0.90$ & $7.45$ & $67.69$ \\
        ChPrune-R18-4x & ${0.44}$ & ${0.45}$ & $3.69$ & $61.56$ \\
        % WeightSVD-R18-4x & $0.45$ & ${0.45}$ & ${2.10}$ & $60.79$ \\

        \arrayrulecolor{black}\bottomrule
    \end{tabular}
    \begin{flushright}
    Entries are shortened, e.g.\ `Channel Pruning' as `ChPrune'. Results for \cite{he2017,zhang2016accelerating} are obtained from \cite{kuzmin2019taxonomy}.
    \end{flushright}
    \label{table:imagenet_resnet}
    \end{minipage}
    \begin{minipage}{\linewidth}
        \small
        \centering
        \caption{Results: EfficientNet on ImageNet}
        \begin{tabular}{lcccc}
            \toprule
            \shortstack{Architecture} & \shortstack{Adds\\($\times 10^6$)} & \shortstack{Mults\\($\times 10^6$)} & \shortstack{\footnotesize{Params}\\($\times 10^6$)} & \shortstack{Acc. \\ (in \%)}\\
            \midrule
            \midrule
            EfficientNet-B1 \cite{efficientnet} & $0.70$ & $0.70$ & $7.80$ & ${78.50}^1$\\
            \textbf{Struct-EffNet (ours)} & ${0.62}$ & ${0.45}$ & ${4.93}$ & $76.40^1$ \\
            % \hdashline
            \arrayrulecolor{black!30}\midrule
            EfficientNet-B0\ \cite{efficientnet} & $0.39$ & $0.39$ & $5.30$ & $76.10^1$ \\
            \arrayrulecolor{black}\bottomrule 
        \end{tabular}
        \label{table:imagenet_efficientnet}
        \end{minipage}
    \end{minipage}
    % \yb{I am revamping these tables right now.. Please ignore for a few minutes :)} \fp{I removed bold in the table as it does not work to our advantage necessarily} \yb{yes, i agree, looks better now}
\end{table}
\endgroup

\subsection{Image Classification}

We present results for ResNets \cite{resnet} in Tables \ref{table:cifar_resnet} and \ref{table:imagenet_resnet}. To demonstrate the efficacy of our method on modern networks, we also show results on MobileNetV2 \cite{mobilenetv2} and EfficientNet\footnote{Our Efficientnet reproduction of baselines (B0 and B1) give results slightly inferior to \cite{efficientnet}. Our Struct-EffNet is created on top of this EfficientNet-B1 baseline.} \cite{efficientnet} in Table~\ref{table:imagenet_mv2} and \ref{table:imagenet_efficientnet}.

To provide a comprehensive analysis, 
%we demonstrate \textit{structured} versions "A" and "B" for each baseline 
% architecture 
% aimed at producing competitive accuracy and drastic compression, respectively. 
for each baseline architecture, we present \textit{structured} counterparts, with version "A" designed to deliver similar accuracies and version "B" for extreme compression ratios. Using different $\{c,n\}$ configurations per-layer, we obtain structured versions with varying levels of reduction in model size and multiplications/additions (please see Supplementary material for details). For the "A" versions of ResNet, we set the compression ratio ($CN^2/cn^2$) to be 2$\times$ for all layers. For the "B" versions of ResNets, we use nonuniform compression ratios per layer. Specifically, we compress stages 3 and 4 drastically (4$\times$) and stages 1 and 2 by $2\times$. Since MobileNet is already a compact model, we design its "A" version to be $1.33\times$ smaller and "B" version to be $2\times$ smaller.

We note that, on low-level hardware, additions are much power-efficient and faster than multiplications \cite{chen2019addernet,horowitz20141}. Since the actual inference time depends on how software optimizations and scheduling are implemented, for most objective conclusions, we provide the number of additions / multiplications and model sizes. Considering observations for sum-pooling on dedicated hardware units \cite{young2018performing}, our structured convolutions can be easily adapted for memory and compute limited devices. %to provide a significant improvement on the power and latency savings.

Compared to the baseline models, the Struct-A versions of ResNets are $2\times$ smaller, while maintaining less than $0.65\%$ loss in accuracy. The more aggressive Struct-B ResNets achieve $60$-$70\%$ model size reduction with about $2$-$3\%$ accuracy drop. Compared to other methods, Struct-56-A is $0.75\%$ better than AMC-Res56 \cite{amc} of similar complexity and Struct-20-A exceeds ShiftResNet20-6 \cite{wu2018shift} by $0.45\%$ while being significantly smaller. Similar trends are observed with Struct-Res18 and Struct-Res50 on ImageNet. Struct-56-A and Struct-50-A achieve competitive performance as compared to the recent GhostNets \cite{ghostnet}. 
% Structured Convolutions can potentially be used in conjunction with other SOTA methods \cite{ghostnet,jaderberg2014speeding} to develop even more efficient models. 
For MobileNetV2 which is already designed to be efficient, Struct-MV2-A achieves further reduction in multiplications and model size with SOTA performance compared to other methods, see Table \ref{table:imagenet_mv2}. Applying structured convolutions to EfficientNet-B1 results in Struct-EffNet that has comparable performance to EfficientNet-B0, as can be seen in Table \ref{table:imagenet_efficientnet}.
%(lesser parameters but more multiplications at a slightly better accuracy).

\begin{figure}[t]
    \centering
    \begin{minipage}{0.25\textwidth}
        \centering
        \includegraphics[width=0.98\textwidth]{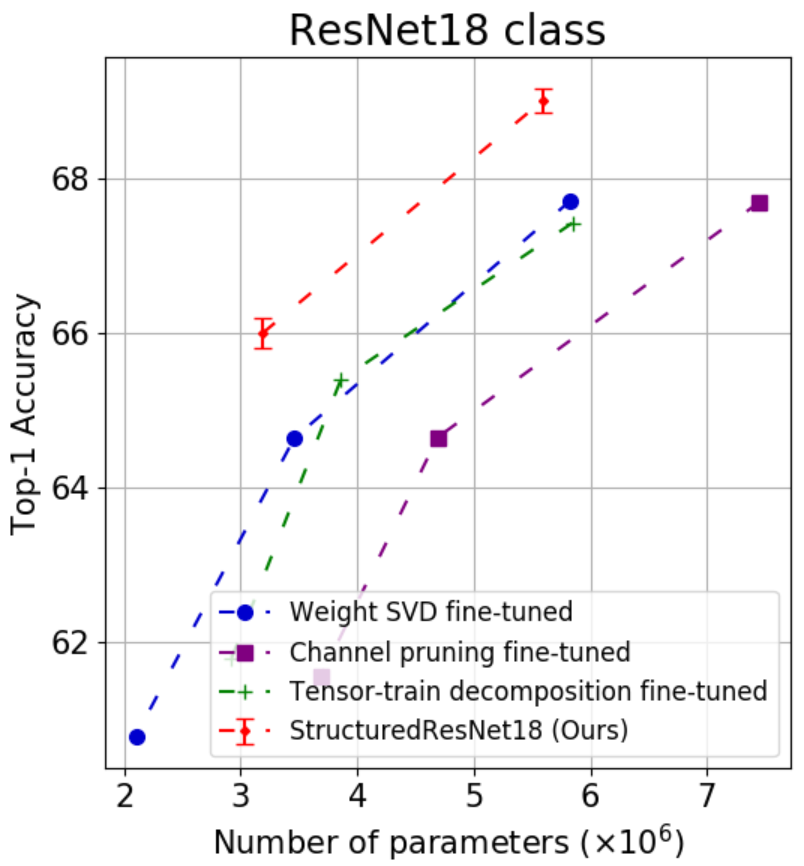}
        % first figure itself

    \end{minipage}\hfill
    % \begin{minipage}{0.33\textwidth}
    %     \centering
    %     \includegraphics[width=0.98\textwidth,height=0.85\textwidth]{Figures/Resnet50_SOTA_parameters.PNG}
    %     % first figure itself
    % \end{minipage}%\hfill
    \centering
    \begin{minipage}{0.25\textwidth}
        \centering
        \includegraphics[width=0.98\textwidth]{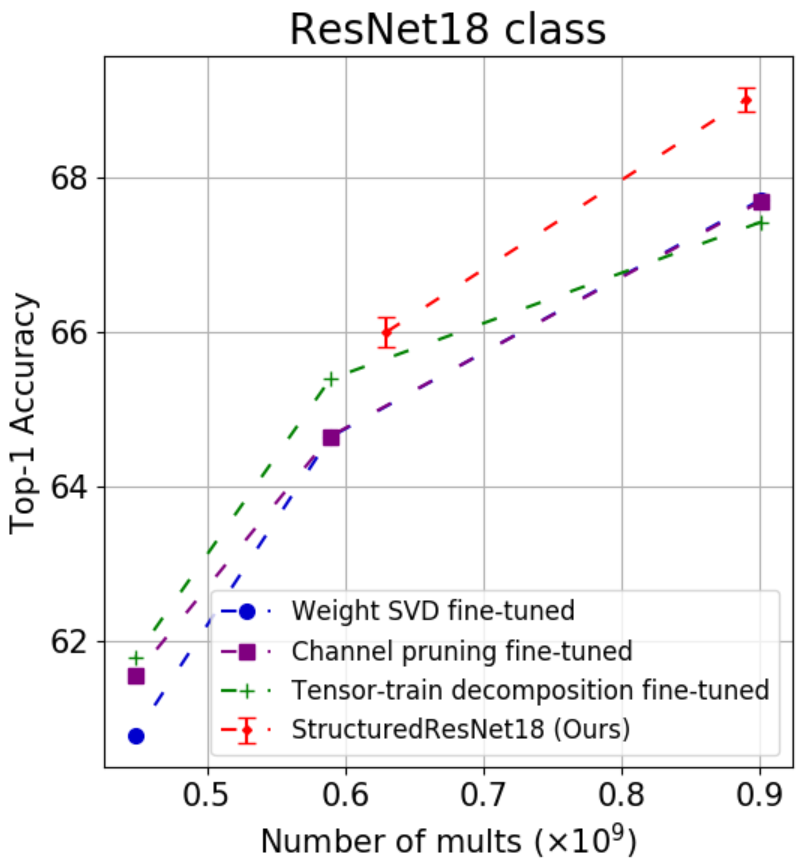}
        % first figure itself

    \end{minipage}\hfill
    \begin{minipage}{0.25\textwidth}
        \centering
        \includegraphics[width=0.98\textwidth]{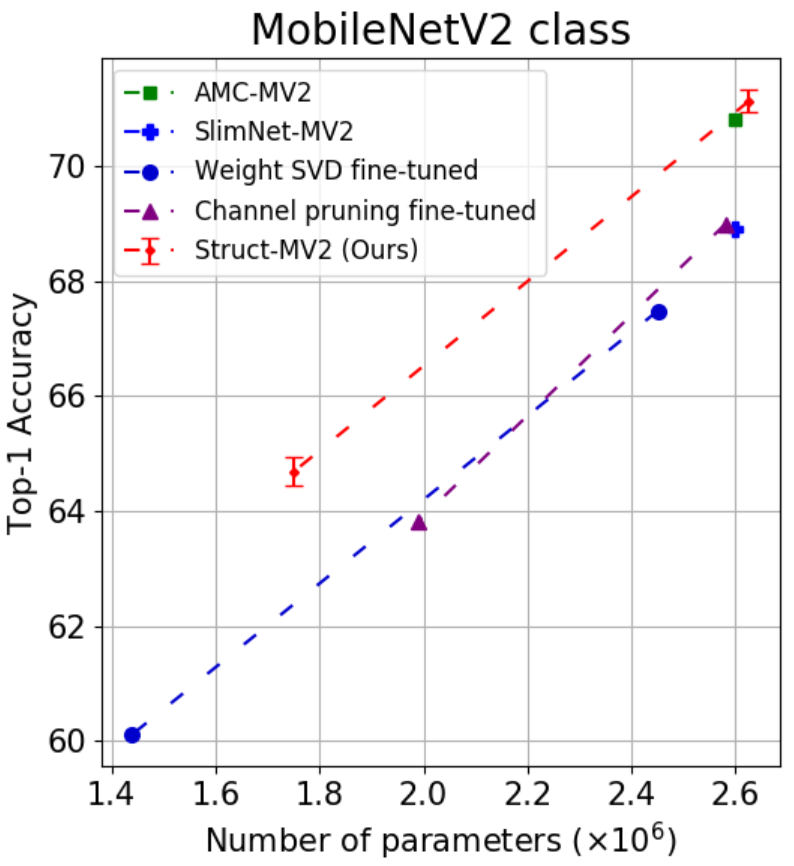}
        % first figure itself
    \end{minipage}\hfill
    % \begin{minipage}{0.33\textwidth}
    %     \centering
    %     \includegraphics[width=0.98\textwidth]{Figures/Resnet50_SOTA_multiplications_wo_spatialSVD.PNG}
    %     % first figure itself
    % \end{minipage}\hfill
    \begin{minipage}{0.25\textwidth}
        \centering
        \includegraphics[width=0.98\textwidth]{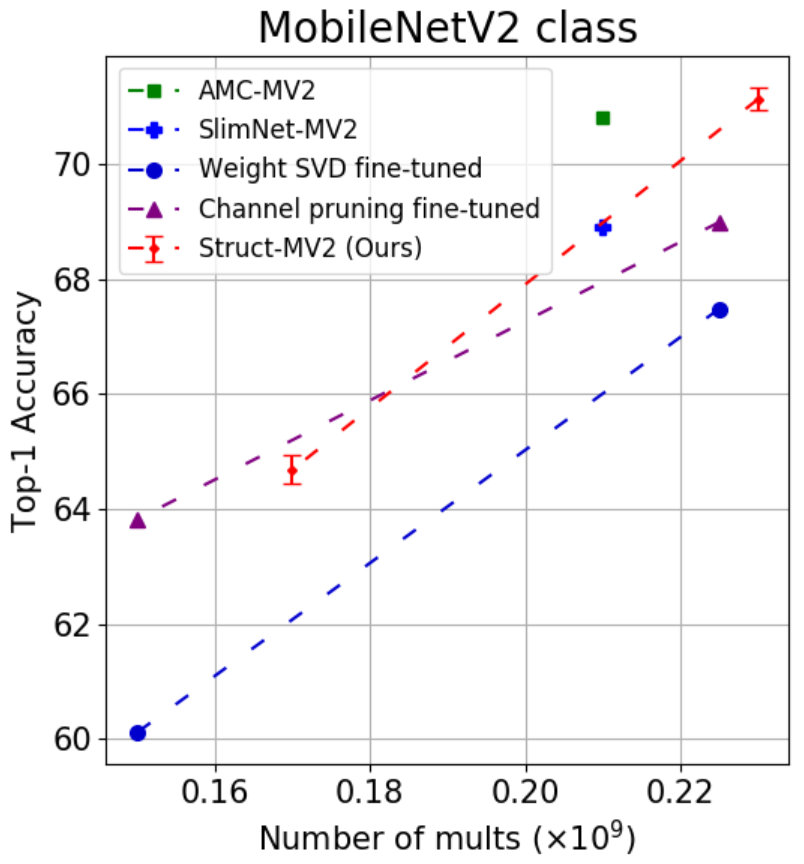}
        % first figure itself
    \end{minipage}
    
    \caption{Comparison with structured compression methods shows `ImageNet top-1 vs model size' trade-off as well as `ImageNet top-1 vs multiplications' trade-off. Note that, we plot the \textbf{mean} $\pm$ \textbf{stddev} (error bars) values for our method, whereas the compared methods have provided their ‘\textbf{best}’ numbers. Also, parameter count for AMC-MV2 is interpolated assuming uniform compression; the actual model would be larger than shown given that AMC \protect\cite{amc} optimizes for MAC count.}
    \label{fig:structured_comparisons}
\end{figure}

The ResNet Struct-A versions have similar number of adds and multiplies (except ResNet50) because, as noted in Sec.~\ref{sec:num_ops}, the sum-pooling contribution is amortized. But sum-pooling starts dominating as the compression gets more aggressive, as can be seen in the number of adds for Struct-B versions. Notably, both "A" and "B" versions of MobileNetV2 observe a dominance of the sum-pooling component. This is because the number of output channels are not enough to amortize the sum-pooling component resulting from the decomposition of the pointwise ($1\!\times\!1$ conv) layers.

Fig.~\ref{fig:structured_comparisons} compares our method with state-of-the-art structured compression methods - WeightSVD \cite{zhang2016accelerating}, Channel Pruning \cite{he2017}, and Tensor-train \cite{su2018tensorized}. Note, the results were obtained from \cite{kuzmin2019taxonomy}.
%which extends these methods with a greedy selection method for per-layer compression ratios.
% \footnote{We were only able to obtain the Tensor-train results for ResNet18 for number of multiplications comparison.}  
Our proposed method achieves approximately $1\%$ improvement over the second best method for ResNet18 ($2\times$) and MobileNetV2 ($1.3\times$). Especially for MobileNetV2, this improvement is valuable since it significantly outperforms all the other methods (see Struct-V2-A in Table \ref{table:imagenet_mv2}).

\subsection{Semantic Segmentation}
\vspace{-15pt}
\begingroup
\begin{table}[h]
    \caption{Evaluation of proposed method on Cityscapes \cite{cordts2016cityscapes} using HRNetV2.}
  % \begin{minipage}{.5\linewidth}
    \small
    \centering
    \begin{tabular}{ccccc}
        \toprule
        \shortstack{HRNetV2-W18\\-Small-v2} & \shortstack{\#adds\\($\times 10^9$)} & \shortstack{\#mults\\($\times 10^9$)} & \shortstack{\#params\\($\times 10^6$)} & \shortstack{Mean IoU \\ (in \%)}\\
        \midrule
        Original & $76.8$ & $77.1$ & $3.9$ & $76.1$ \\
        Struct-HR-A & $54.3$ & $54.0$ & $1.9$ & $74.6$ \\
        % Struct-HR-B & $60.7$ & $48.4$ & $1.2$ & $--$ \\
        \bottomrule 
    \end{tabular}
    \label{table:hrnet}
    \label{table:cityscapes}
\end{table}
\endgroup
\vspace{-5pt}
After demonstrating the superiority of our method on image classification, we evaluate it for semantic segmentation that requires reproducing fine details around object boundaries. We apply our method to a recently developed state-of-the-art HRNet~\cite{WangSCJDZLMTWLX19}. Table \ref{table:cityscapes} shows that the structured convolutions can significantly improve our segmentation model efficiency: HRNet model is reduced by 50\% in size, and 30\% in number of additions and multiplications, while having only 1.5\% drop in mIoU. More results for semantic segmentation can be found in the supplementary material.

% \vspace{-0.1in}

%The Cityscapes dataset~\cite{cordts2016cityscapes} contains 19 classes of objects, e.g., person, building, traffic sign, etc.\ and the segmentation task is to provide a pixel-level labeling for the raw image input (1024$\smalltimes$2048). We apply our structured convolution method to a recent efficient model HRNet~\cite{WangSCJDZLMTWLX19} and a widely known model PSPNet~\cite{zhao2017pyramid}. In Table \ref{table:cityscapes}, we show that our proposed structured convolution effectively reduces the model size, as well as number of addition and multiplication. Note that segmentation problem requires a network (mostly its convolution layers) to reproduce all the fine details of objects' boundaries, which leads segmentation network being more difficult to be structured and compressed.  

% \vspace{-0.1in}

\subsection{Computational overhead of Structural Regularization term}
The proposed method involves computing the Structural Regularization term (\ref{eq:structural_regularization_loss}) during training. Although the focus of this work is more on inference efficiency, we measure the memory and time-per-iteration for training \textit{with} and \textit{w/o} the Structural Regularization loss on an NVIDIA V100 GPU. We report these numbers for Struct-18-A and Struct-MV2-A architectures below. As observed, the additional computational cost of the regularization term is negligible for a batchsize of $256$. This is because, mathematically, the regularization term, $\sum_{l=1}^L\frac{\norm{(I-A_lA_l^+)W_l}_F}{\norm{W_l}_F}$, is independent of the input size. Hence, when using a large batchsize for training, the regluarization term's memory and runtime overhead is relatively small (less than $5\%$ for Struct-MV2-A and $10\%$ for Struct-18-A).

% The size of the $A_l$ matrix depends on the convolution kernel size. So, for architectures with larger convolution kernels, the cost of regularization term will be proportionally higher. But at the same time, the cost of computing $\mathcal{L}_{task}$ also increases

% Although the focus of this work is more on inference efficiency, we measure memory and time-per-iteration for training \textit{with} and \textit{w/o} Structural Regularization loss, on an NVIDIA V100 GPU. Mathematically, the regularization term, $\sum_{l=1}^L\frac{\norm{(I-A_lA_l^+)W_l}_F}{\norm{W_l}_F}$, is independent of the input size. Hence, when using a large batchsize, the regluarization term's memory and runtime overhead is relatively small as shown in the table.

% \begin{wraptable}[5]{R}{0.5\textwidth}
\begin{table}[h]
	\caption{Memory and runtime costs of training \textit{with} and \textit{w/o} Structural Regularization (SR) loss}
    \small
	\centering
    \small
    % \vspace{-12pt}
	\begin{tabular}{l  c  c  c c}
        \toprule
        	\multirow{2}{*}{\shortstack{Training costs\\ ({batchsize=256})}} & \multicolumn{2}{c}{Struct-18-A} & \multicolumn{2}{c}{Struct-MV2-A} \\ 
        % 	\multicolumn{2}{c}{Struct-EffNet} \\
        	\cmidrule{2-5}
        	& {Mem} & {seconds / iter} & {Mem} & {seconds / iter} \\ % & {GPU Mem} & {secs / iter} \\
			
			\midrule
	
			With SR loss & $9.9$GB & $0.46$s & $18.8$GB & $0.38$s \\ % & $35.5$GB & $1.04$s \\
			Without SR loss & $9.2$GB & $0.44$s & $17.9$GB & $0.37$s \\ % & $34.2$GB & $1.01$s \\
        \bottomrule
	\end{tabular}
	\label{table:training_cost}
\end{table}
% \end{wraptable}

\subsection{Directly training Structured Convolutions as an architectural feature}
In our proposed method, we train the architecture with original $C\!\times\!N\!\times\!N$ kernels in place and the regularization loss imposes desired structure on these kernels. At the end of training, we decompose the $C\!\times\!N\!\times\!N$ kernels and replace each with a sum-pooling layer and smaller layer of $c\!\times\!n\!\times\!n$ kernels.

A more \textit{direct} approach could be to train the decomposed architecture (with the sum-pooling + $c\!\times\!n\!\times\!n$ layers in place) directly. The regularization term is not required in this direct approach, as there is no decomposition step, hence eliminating the computation overhead shown in Table \ref{table:training_cost}. We experimented with this direct training method and observed that the regularization based approach always outperformed the direct approach (by $1.5$\% for Struct-18-A and by $0.9$\% for Struct-MV2-A). This is because, as pointed out in Sec.~\ref{sec:regularization}, the direct method optimizes the weights in a restricted subspace of $c\!\times\! n\!\times\! n$ kernels right from the start, whereas the regularization based approach gradually moves the weights from the larger ($C\!\times\!N\!\times\!N$) subspace to the restricted subspace with gradual imposition of the structure constraints via the regularization loss.

% For better understanding, we provide pseudocodes for the direct and proposed approaches in the apendix.

\section{Conclusion}

In this work, we propose Composite Kernels and Structured Convolutions in an attempt to exploit redundancy in the \textit{implicit} structure of convolution kernels. We show that Structured Convolutions can be decomposed into a computationally cheap sum-pooling component followed by a significantly smaller convolution, by training the model using an intuitive structural regularization loss. The effectiveness of the proposed method is demonstrated via extensive experiments on image classification and semantic segmentation benchmarks. Sum-pooling relies purely on additions, which are known to be extremely power-efficient. Hence, our method shows promise in deploying deep models on low-power devices. Since our method keeps the convolutional structures, it allows integration of further model compression schemes, which we leave as future work.

\section*{Broader Impact}

The method proposed in this paper promotes the adaption of deep learning neural networks into memory and compute limited devices, allowing a broader acceptance of machine learning solutions for a spectrum of real-life use cases. By reducing the associated hardware costs of the neural network systems, it aims at making such technology affordable to larger communities. It empowers people by facilitating access to the latest developments in this discipline of science. It neither leverages biases in data nor demands user consent for the use of data. 

%Authors are required to include a statement of the broader impact of their work, including its ethical aspects and future societal consequences. Authors should discuss both positive and negative outcomes, if any. For instance, authors should discuss a) who may benefit from this research, b) who may be put at disadvantage from this research, c) what are the consequences of failure of the system, and d) whether the task/method leverages biases in the data. If authors believe this is not applicable to them, authors can simply state this. Use unnumbered first level headings for this section, which should go at the end of the paper. 

\section*{Acknowledgements}
We would like to thank our Qualcomm AI Research colleagues for their support and assistance, in particular that of Andrey Kuzmin, Tianyu Jiang, Khoi Nguyen, Kwanghoon An and Saurabh Pitre.
% \begin{ack}
% Use unnumbered first level headings for the acknowledgments. All acknowledgments go at the end of the paper before the list of references. Moreover, you are required to declare funding (financial activities supporting the submitted work) and competing interests (related financial activities outside the submitted work). More information about this disclosure can be found at: \url{https://neurips.cc/Conferences/2020/PaperInformation/FundingDisclosure}.

% Do {\bf not} include this section in the anonymized submission, only in the final paper. You can use the \texttt{ack} environment provided in the style file to autmoatically hide this section in the anonymized submission.
% \end{ack}

\bibliographystyle{plain}
\bibliography{structured}

\newpage

\appendix
\section{Appendix}

\subsection{Structured Convolutions with arbitrary {Padding, Stride} and {Dilation}}

In the main paper, we showed that a Structured Convolution can be decomposed into a Sum-Pooling component followed by a smaller convolution operation with a kernel composed of the $\alpha$'s.
In this section, we discuss how to calculate the equivalent stride, padding and dilation needed for the resulting decomposed sum-pooling and convolution operations.

\subsubsection{Padding}
\begin{figure}[h]
    \centering
    \includegraphics[width=0.7\linewidth]{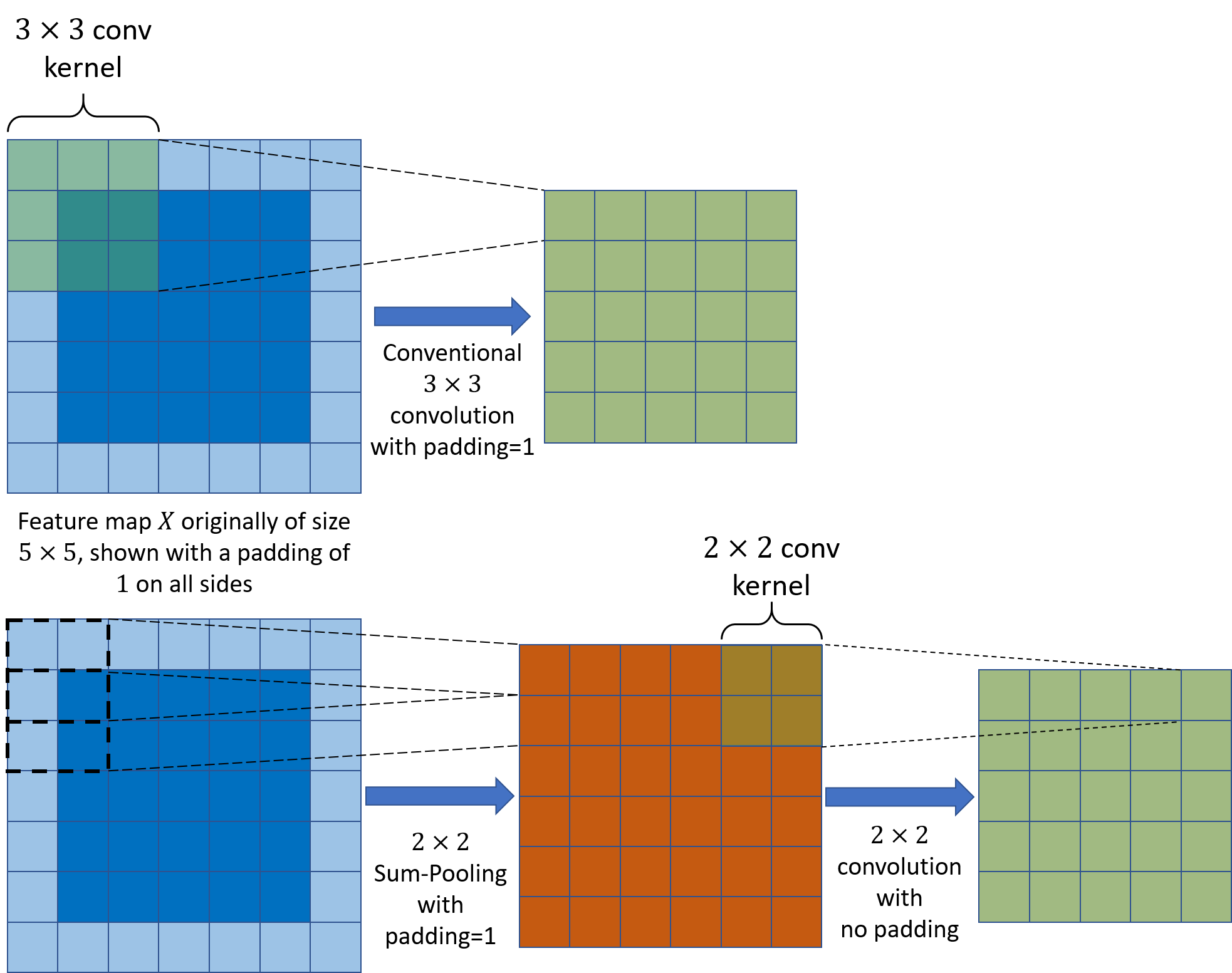}
    \caption{Decomposition of a $3\times 3$ Structured Convolution with a padding of $1$. Top shows the conventional operation of the convolution. Bottom shows the equivalent operation using sum-pooling.}
    \label{fig:padding_illustration}
\end{figure}

The easiest of these three attributes is \textit{padding}. Fig.~\ref{fig:padding_illustration} shows an example of a structured convolution with a $3\times 3$ kernel (i.e. $N=3$) with underlying parameter $n=2$. Hence, it can be decomposed into a $2\times 2$ sum-pooling operation followed by a $2\times 2$ convolution. As shown in the figure, to preserve the same output after the decomposition, the sum-pooling component should use the same padding as the \textit{original} $3\times 3$ convolution, whereas the smaller $2\times 2$ convolution is performed without padding. 

This leads us to a more general result that - if the original convolution uses a padding of $p$, then, after the decomposition, the sum-pooling should be performed with padding $p$ and the smaller convolution (with $\alpha$'s) should be performed without padding.

\subsubsection{Stride}
The above rule can be simply extended to the case where the original $3\times 3$ structured convolution has a stride associated with it. The general rule is - if the original convolution uses a stride of $s$, then, after the decomposition, the sum-pooling should be performed with a stride of $1$ and the smaller convolution (with $\alpha$'s) should be performed with a stride of $s$.

\subsubsection{Dilation}
\begin{figure}
    \centering
    \includegraphics[width=0.7\linewidth]{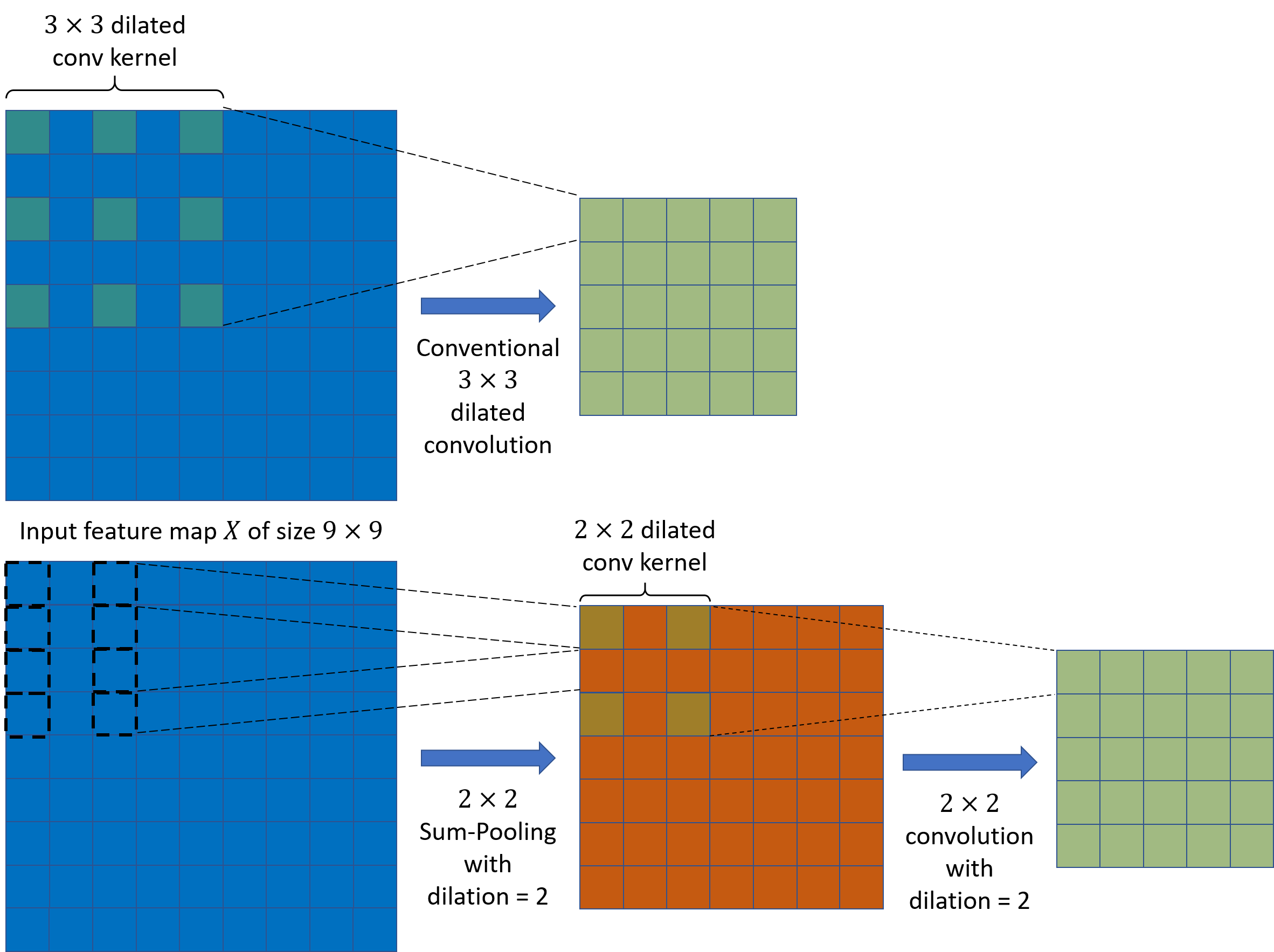}
    \caption{Decomposition of a $3\times 3$ Structured Convolution with a dilation of $2$. Top shows the conventional operation of the convolution. Bottom shows the equivalent operation using sum-pooling.}
    \label{fig:dilation_illustration}
\end{figure}

Dilated or atrous convolutions are prominent in semantic segmentation architectures. Hence, it is important to consider how we can decompose dilated structured convolutions. Fig.~\ref{fig:dilation_illustration} shows an example of a $3\times 3$ structured convolution with a dilation of $2$. As can be seen in the figure, to preserve the same output after decomposition, \textit{both} the sum-pooling component and the smaller convolution (with $\alpha$'s) has to be performed with a dilation factor same as the original convolution.

\begin{figure}
    \centering
    \includegraphics[width=0.65\linewidth]{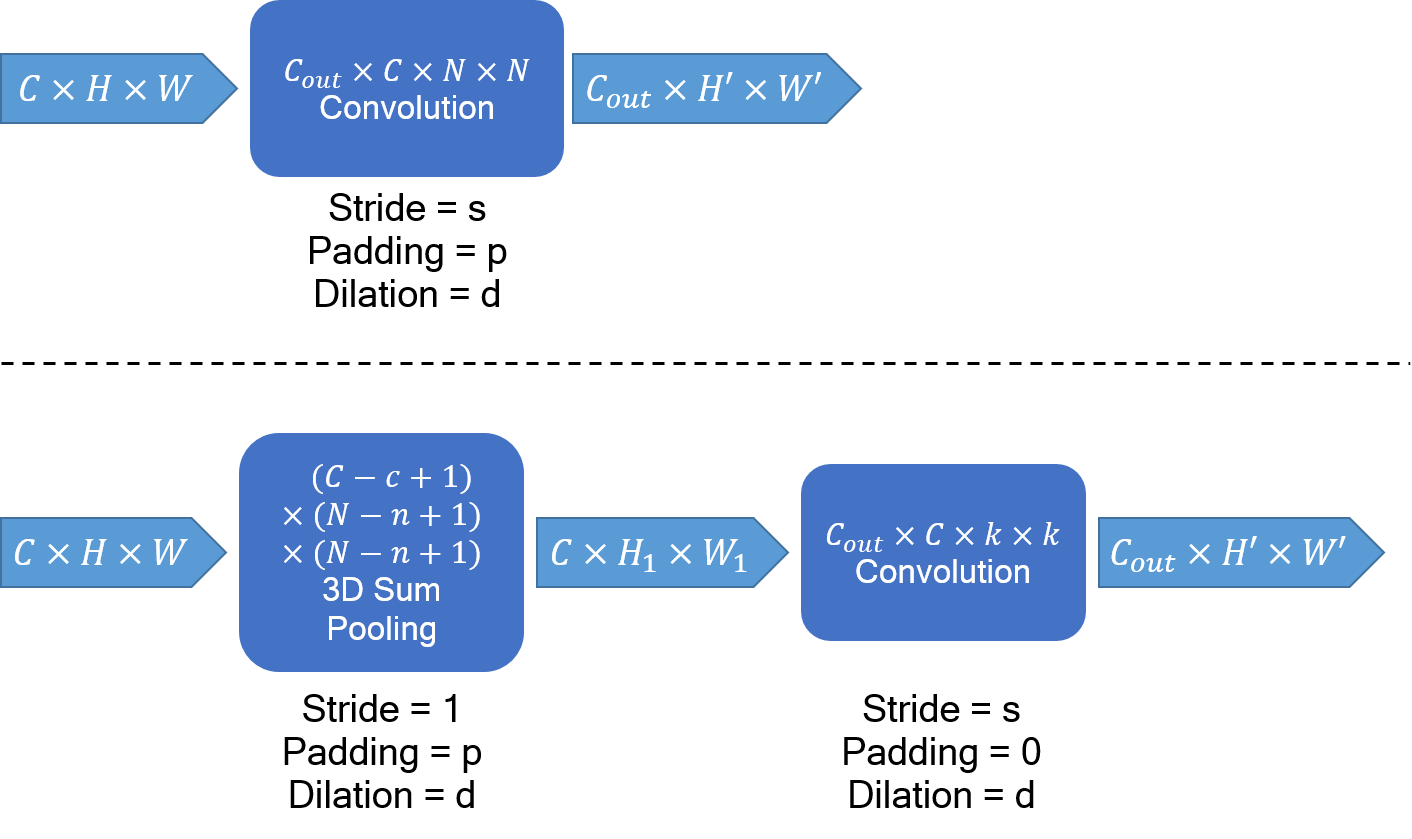}
    \caption{Decomposition of a general structured convolution with stride, padding and dilation. The blocked arrows indicate the dimensions of the input and output tensors. Top shows the conventional operation of the convolution. Bottom shows the equivalent operation using sum-pooling.}
    \label{fig:general_decomposition}
\end{figure}

Fig.~\ref{fig:general_decomposition} summarizes the aforementioned rules regarding padding, stride and dilation. 
% The reader can easily verify these rules with a few lines of code in PyTorch.

\subsection{Training Implementation Details}
\textbf{Image Classification. } For both ImageNet and CIFAR-10 benchmarks, we train all the ResNet architectures from scratch with the Structural Regularization (SR) loss. We set $\lambda$ to $0.1$ for the Struct-A versions and $1.0$ for the Struct-B versions throughout training. For MobileNetV2, we first train the deep network from scratch \textit{without} SR loss (i.e. $\lambda=0$) for $90$ epochs to obtain pretrained weights and then apply SR loss with $\lambda=1.0$ for further $150$ epochs. For EfficientNet-B0, we first train without SR loss for $90$ epochs and then apply SR loss with $\lambda=1.0$ for further $250$ epochs.

For CIFAR-10, we train the ResNets for $200$ epochs using a batch size of $128$ and an initial learing rate of $0.1$ which is decayed by a factor of $10$ at $80$ and $120$ epochs. We use a weight decay of $0.0001$ throughout training. On ImageNet, we use a cosine learning rate schedule with an SGD optimizer for training all architectures. We train the ResNets using a batch size of $256$ and weight decay of $0.0001$ for $200$ epochs starting with an initial learning rate of $0.1$. 

For MobileNetV2, we use a weight decay of $0.00004$ and batch size $128$ throughout training. In the first phase (with $\lambda=0$), we use an initial learning rate of $0.5$ for $90$ epochs and in the second phase, we start a new cosine schedule with an initial learning rate of $0.1$ for the next $150$ epochs. We train EfficientNet-B0 using Autoaugment, a weight decay of $0.00004$ and batch size $384$. We use an initial learning rate of $0.5$ in the first phase and we start a new cosine schedule for the second phase with an initial learning rate of $0.1$ for the next $250$ epochs.

\textbf{Semantic Segmentation. } 
For training Struct-HRNet-A on Cityscapes, we start from a pre-trained HRNet model and train using structural regularization loss. We set $\lambda$ to $1.0$. We use a cosine learning rate schedule with an initial learning rate of $0.01$. The use image resolution of $1024 \times 2048$ for training, same as the original image size. We train for 90000 iterations using a batch size of 4. 

We show additional results with PSPNet in Sec.~\ref{sec:additional_results} below. We follow a similar training process for training Struct-PSPNet-A where we start from a pre-trained PSPNet101 \cite{zhao2017pyramid}.

\subsection{Additional results on Semantic Segmentation}
\label{sec:additional_results}
In Table \ref{table:hrnet2} and \ref{table:pspnet}, we present additional results for HRNetV2-W18-Small-v1 \cite{WangSCJDZLMTWLX19} (note this is different from HRNetV2-W18-Small-v2 reported in the main paper) and PSPNet101 \cite{zhao2017pyramid} on Cityscapes dataset. 
\begin{table}[h]
    \caption{Evaluation of proposed method on Cityscapes using HRNetV2-W18-Small-v1 \cite{WangSCJDZLMTWLX19}.}
  % \begin{minipage}{.5\linewidth}
    \small
    \centering
    \begin{tabular}{ccccc}
        \toprule
        \shortstack{HRNetV2-W18\\-Small-v1} & \shortstack{\#adds\\($\times 10^9$)} & \shortstack{\#mults\\($\times 10^9$)} & \shortstack{\#params\\($\times 10^6$)} & \shortstack{mIoU \\ (in \%)}\\
        \midrule
        Original & $33.4$ & $33.6$ & $1.5$ & $70.3$ \\
        Struct-HR-A-V1 & $24.6$ & $24.5$ & $0.77$ & $67.5$ \\
        % Struct-HR-B & $60.7$ & $48.4$ & $1.2$ & $--$ \\
        \bottomrule 
    \end{tabular}
    \label{table:hrnet2}
    \caption{Evaluation of proposed method on Cityscapes using PSPNet101 \cite{zhao2017pyramid}.}
    \small
    \centering
    \begin{tabular}{ccccc}
        \toprule
        \shortstack{PSPNet101} & \shortstack{\#adds\\($\times 10^9$)} & \shortstack{\#mults\\($\times 10^9$)} & \shortstack{\#params\\($\times 10^6$)} & \shortstack{mIoU \\ (in \%)}\\
        \midrule
        Original & $2094$ & $2096$ & $68.1$ & $79.3$ \\
        Struct-PSP-A & $1325$ & $1327$ & $43.0$ & 76.6 \\
        \bottomrule 
    \end{tabular}
    \label{table:pspnet}
\end{table}

\subsection{Layer-wise compression ratios for compared architectures}
As mentioned in the Experiments section of the main paper, we use non-uniform selection for the per-layer compression ratios ($CN^2/cn^2$) for MobileNetV2 and EfficientNet-B0 as well as HRNet for semantic segmentation. Tables \ref{table:struct-mv2-a} and \ref{table:struct-mv2-b} show the layerwise $\{c,n\}$ parameters for each layer of the Struct-MV2-A and Struct-MV2-B architectures. Table \ref{table:struct-effnet} shows these per-layer $\{c,n\}$ parameters for Struct-EffNet.

For Struct-HRNet-A, we apply Structured Convolutions only in the spatial dimension, i.e. we use $c=C$, hence there's no decomposition across the channel dimension. For $3\smalltimes 3$ convolutional kernels, we use $n=2$, which means a $3 \smalltimes 3$ convolution is decomposed into a $2\smalltimes2$ sum pooling followed by a $2 \smalltimes 2$ convolution. And for $1\smalltimes 1$ convolutions, where $N=1$, we use $n=1$ which is the only possiblility for $n$ since $1\leq n\leq N$. We do not use Structured Convolutions in the initial two convolution layers and last convolution layer. 

For Struct-PSPNet, similar to Struct-HRNet-A, we apply use structured convolutions in all the convolution layers except the first and last layer. For $3 \smalltimes 3$ convolutions, the structured convolution uses $c=C$ and $n=2$. For $1 \smalltimes 1$ convolutions, the structured convolution uses $c=round(2 \smalltimes C/3)$ and $n=1$.

% \section{Comparison with Alternative Training Method}

\subsection{Sensitivity of Structural Regularization w.r.t \texorpdfstring{$\lambda$}{lambda}}
In Sec.~5.1, we introduced the Structural Regularization (SR) loss and proposed to train the network using this regularization with a weight $\lambda$. In this section, we investigate the variation in the final performance of the model (after decomposition) when trained with different values of $\lambda$. 

We trained Struct-Res18-A and Struct-Res18-B with different values of $\lambda$. Note that when training both "A" and "B" versions, we start with the original architecture for ResNet18 and train it from scratch with the SR loss. After this first step, we then decompose the weights using $\alpha_l=A^+W_l$ to get the decomposed architecture. Tables \ref{table:lambda_sensitivity_A} and \ref{table:lambda_sensitivity_B} show the accuracy of Struct-Res18-A and Struct-Res18-B both pre-decomposition and post-decomposition.

\begin{table}[h]
    \parbox{.48\linewidth}{
	\caption{ImageNet performance of Struct-Res18-A trained with different $\lambda$}
	\small
	\centering
	\begin{tabular}{c  c c }
        \toprule
        	$\lambda$ & \shortstack{Acc. (before \\decomposition)} & \shortstack{Top-1 Acc. (after \\decomposition)} \\
			\midrule
			$1.0$ & $69.08\%$ & $69.11\%$ \\
			$0.5$ & $69.21\%$ & $69.09\%$ \\
			$0.1$ & $69.17\%$ & $\mathbf{69.13}\%$ \\
			$0.05$ & $69.16\%$ & $69.05\%$ \\
			$0.01$ & $69.20\%$ & $69.09\%$ \\
			$0.001$ & ${69.23}\%$ & $68.57\%$ \\
        \bottomrule
	\end{tabular}
	\label{table:lambda_sensitivity_A}
	}
	\hfill
    \parbox{.48\linewidth}{
	\caption{ImageNet performance of Struct-Res18-B trained with different $\lambda$}
	\small
	\centering
	\begin{tabular}{c  c c }
        \toprule
        	$\lambda$ & \shortstack{Acc. (before \\decomposition)} & \shortstack{Top-1 Acc. (after \\decomposition)} \\
			\midrule
			$1.0$ & $66.04\%$ & $\mathbf{66.19}\%$ \\
			$0.5$ & ${66.11}\%$ & $66.01\%$ \\
			$0.1$ & $66.01\%$ & ${65.89}\%$ \\
			$0.05$ & $65.97\%$ & $65.65\%$ \\
			$0.01$ & $65.99\%$ & $64.47\%$ \\
			$0.001$ & ${65.91}\%$ & $58.19\%$ \\
        \bottomrule
	\end{tabular}
	\label{table:lambda_sensitivity_B}
	}
\end{table}

\begin{table}[h]
	\caption{ImageNet performance \textit{before} and \textit{after} decomposition for other architectures}
	\small
	\centering
	\begin{tabular}{c  c c }
        \toprule
        	Architecture & \shortstack{Acc. (before \\decomposition)} & \shortstack{Top-1 Acc. (after \\decomposition)} \\
			\midrule
			Struct-50-A & $75.68\%$ & $75.65\%$ \\
			Struct-50-B & $73.52\%$ & $73.41\%$ \\
			\midrule
			Struct-V2-A & $71.33\%$ & $71.29\%$ \\
			Struct-V2-B & $65.01\%$ & $64.93\%$ \\
			\midrule
			Struct-EffNet & $76.59\%$ & $76.40\%$ \\
		\bottomrule
	\end{tabular}
	\label{table:lambda_sensitivity_others}
\end{table}

From Table \ref{table:lambda_sensitivity_A}, we can see that the accuracy after decomposition isn't affected much by the choice of $\lambda$. When $\lambda$ varies from $0.01$ to $1.0$, the post-decomposition accuracy only changed by $0.08\%$. Similar trends are observed in Table \ref{table:lambda_sensitivity_B} when we are compressing more aggressively. But the sensitivity of the performance w.r.t.~$\lambda$ is slightly higher in the "B" version. Also, we can see that when $\lambda=0.001$, the difference between pre-decomposition and post-decomposition accuracy is significant. Since $\lambda$ is very small in this case, the Structural Regularization loss does not impose the desired structure on the convolution kernels effectively. As a result, after decomposition, it leads to a loss in accuracy.

In Table \ref{table:lambda_sensitivity_others}, we show the ImageNet performance of other architectures (from Tables 1, 2, 3, 4 of main paper) before and after the decomposition is applied.

\subsection{Expressive power of the Sum-Pooling component}
To show that the sum-pooling layers indeed capture meaningful features, we perform a toy experiment where we swap all $3\times 3$ depthwise convolution kernels in MobileNetV2 with $2\times 2$ kernels and train the architecture. We observed that this leads to a severe performance degradation of $4.5\%$ compared to the Struct-V2-A counterpart. This, we believe, is due to the loss of receptive field that was being captured by the sum-pooling part of structured convolutions.

\subsection{Inference Latency}
In Sec.\ 6.1 of the paper, we pointed out that the actual inference time of our method depends on how the software optimizations and scheduling are implemented. Additions are much faster and power efficient than multiplications on low-level hardware \cite{chen2019addernet,horowitz20141}. However, this is not exploited on conventional platforms like GPUs which use FMA (Fused Multiply-Add) instructions. Considering hardware accelerator implementations \cite{young2018performing} for sum-pooling, the theoretical gains of structured convolutions can be realized. We provide \textit{estimates} for the latencies based on measurements on a Intel Xeon CPU W-2123 platform assuming that the software optimizations and scheduling for the sum-pooling operation are implemented. Please refer the table below.

% As mentioned in Sec.\ 6.1 of the main paper, additions are much more efficient than multiplications \cite{chen2019addernet,horowitz20141} on low-level hardware. But this benefit is not observed on conventional hardware platforms, e.g. CPUs and especially GPUs which use FMA (fused multiply-add) instructions. Hence, the theoretical speedups reported in Tables 1-4 are not realizable on these platforms. However, we provide \textit{estimates} for the latency observed on a Intel Xeon CPU W-2123 platform assuming that the software optimizations and scheduling for the sum-pooling operation are implemented. Please refer the table below.

\begin{table}[h]
    \small
	\centering
	\begin{tabular}{ c c  c c  c c }
        \toprule
        	 {ResNet18} & {0.039s} & {MobilenetV2} & {0.088s} & {EfficientNet-B1} & {0.114s} \\
			\midrule
            % \specialrule{\cmidrulewidth}{0pt}{0pt}
            Struct-18-A & \textbf{0.030s} & {Struct-MV2-A} & \textbf{0.078s} & {Struct-EffNet} & \textbf{0.101s} \\
        \bottomrule
	\end{tabular}
	\label{table:inference_cost}
\end{table}

% \subsection{Pseudocode for the Proposed approach and Direct training approach (Sec. 6.4)}
% In the Sec. 6.4 of the main paper, we described a "direct" training approach and compared it with our proposed approach with Structural Regularization. For better understanding of the reader, we provide pseudocodes for both below.

% \begin{algorithm}[H]
% \SetAlgoLined
% \KwIn{Training data, Original architecture}
% %  \Initialize{Model weight with }
% Step 1:\;
%  \For{epoch = 1,...,num\_epochs}{

%  }

%  \caption{Proposed Approach}
% \end{algorithm}

% \bibliographystyle{plain}
% \bibliography{structured}

\begin{table}[h]
    \begin{minipage}{.45\linewidth}
    \small
    \centering
    \caption{Layerwise $\{c,n\}$ configuration for Struct-MV2-A architecture}
    \begin{tabular}{cccc}
        \toprule
        Idx & \shortstack{Dimension\\$C_{out}\times C\times N\times N$} & $c$ & $n$ \\
        % % \cmidrule(lr){2-4}
        % & $C_{out}$ & $C$ & $N$ & \\
        \midrule
        1 & $32 \times 3 \times 3 \times 3$ & 3 & 3\\
        2 & $32 \times 1 \times 3 \times 3$ & 1 & 3\\
        3 & $16 \times 32 \times 1 \times 1$ & 32 & 1\\
        4 & $96 \times 16 \times 1 \times 1$ & 16 & 1\\
        5 & $96 \times 1 \times 3 \times 3$ & 1 & 3\\
        6 & $24 \times 96 \times 1 \times 1$ & 96 & 1\\
        7 & $144 \times 24 \times 1 \times 1$ & 24 & 1\\
        8 & $144 \times 1 \times 3 \times 3$ & 1 & 3\\
        9 & $24 \times 144 \times 1 \times 1$ & 144 & 1\\
        10 & $144 \times 24 \times 1 \times 1$ & 24 & 1\\
        11 & $144 \times 1 \times 3 \times 3$ & 1 & 3\\
        12 & $32 \times 144 \times 1 \times 1$ & 144 & 1\\
        13 & $192 \times 32 \times 1 \times 1$ & 32 & 1\\
        14 & $192 \times 1 \times 3 \times 3$ & 1 & 3\\
        15 & $32 \times 192 \times 1 \times 1$ & 192 & 1\\
        16 & $192 \times 32 \times 1 \times 1$ & 32 & 1\\
        17 & $192 \times 1 \times 3 \times 3$ & 1 & 3\\
        18 & $32 \times 192 \times 1 \times 1$ & 192 & 1\\
        19 & $192 \times 32 \times 1 \times 1$ & 32 & 1\\
        20 & $192 \times 1 \times 3 \times 3$ & 1 & 3\\
        21 & $64 \times 192 \times 1 \times 1$ & 192 & 1\\
        22 & $384 \times 64 \times 1 \times 1$ & 64 & 1\\
        23 & $384 \times 1 \times 3 \times 3$ & 1 & 3\\
        24 & $64 \times 384 \times 1 \times 1$ & 384 & 1\\
        25 & $384 \times 64 \times 1 \times 1$ & 64 & 1\\
        26 & $384 \times 1 \times 3 \times 3$ & 1 & 3\\
        27 & $64 \times 384 \times 1 \times 1$ & 384 & 1\\
        28 & $384 \times 64 \times 1 \times 1$ & 64 & 1\\
        29 & $384 \times 1 \times 3 \times 3$ & 1 & 3\\
        30 & $64 \times 384 \times 1 \times 1$ & 384 & 1\\
        31 & $384 \times 64 \times 1 \times 1$ & 64 & 1\\
        32 & $384 \times 1 \times 3 \times 3$ & 1 & 3\\
        33 & $96 \times 384 \times 1 \times 1$ & 384 & 1\\
        34 & $576 \times 96 \times 1 \times 1$ & 96 & 1\\
        35 & $576 \times 1 \times 3 \times 3$ & 1 & 3\\
        36 & $96 \times 576 \times 1 \times 1$ & 576 & 1\\
        37 & $576 \times 96 \times 1 \times 1$ & 96 & 1\\
        38 & $576 \times 1 \times 3 \times 3$ & 1 & 3\\
        39 & $96 \times 576 \times 1 \times 1$ & 576 & 1\\
        40 & $576 \times 96 \times 1 \times 1$ & 96 & 1\\
        41 & $576 \times 1 \times 3 \times 3$ & 1 & 3\\
        42 & $160 \times 576 \times 1 \times 1$ & 576 & 1\\
        43 & $960 \times 160 \times 1 \times 1$ & 160 & 1\\
        44 & $960 \times 1 \times 3 \times 3$ & 1 & 3\\
        45 & $160 \times 960 \times 1 \times 1$ & 960 & 1\\
        46 & $960 \times 160 \times 1 \times 1$ & 160 & 1\\
        47 & $960 \times 1 \times 3 \times 3$ & 1 & 3\\
        48 & $160 \times 960 \times 1 \times 1$ & 960 & 1\\
        49 & $960 \times 160 \times 1 \times 1$ & 160 & 1\\
        50 & $960 \times 1 \times 3 \times 3$ & 1 & 3\\
        51 & $320 \times 960 \times 1 \times 1$ & 840 & 1\\
        52 & $1280 \times 320 \times 1 \times 1$ & 160 & 1\\
        \scriptsize{classifier} & $1000 \times 1280 \times 1 \times 1$ & 640 & 1\\
        \bottomrule 
    \end{tabular}
    \label{table:struct-mv2-a}
    \end{minipage}
    \hfill
    \begin{minipage}{.45\linewidth}
    \small
    \centering
    \caption{Layerwise $\{c,n\}$ configuration for Struct-MV2-B architecture}
    \begin{tabular}{cccc}
        \toprule
        Idx & \shortstack{Dimension\\$C_{out}\times C\times N\times N$} & $c$ & $n$ \\
        % \cmidrule(lr){2-4}
        % & $C_{out}$ & $C$ & $N$ & \\
        \midrule
        1 & $32 \times 3 \times 3 \times 3$ & 3 & 3\\
        2 & $32 \times 1 \times 3 \times 3$ & 1 & 3\\
        3 & $16 \times 32 \times 1 \times 1$ & 32 & 1\\
        4 & $96 \times 16 \times 1 \times 1$ & 16 & 1\\
        5 & $96 \times 1 \times 3 \times 3$ & 1 & 3\\
        6 & $24 \times 96 \times 1 \times 1$ & 48 & 1\\
        7 & $144 \times 24 \times 1 \times 1$ & 12 & 1\\
        8 & $144 \times 1 \times 3 \times 3$ & 1 & 3\\
        9 & $24 \times 144 \times 1 \times 1$ & 72 & 1\\
        10 & $144 \times 24 \times 1 \times 1$ & 12 & 1\\
        11 & $144 \times 1 \times 3 \times 3$ & 1 & 3\\
        12 & $32 \times 144 \times 1 \times 1$ & 72 & 1\\
        13 & $192 \times 32 \times 1 \times 1$ & 16 & 1\\
        14 & $192 \times 1 \times 3 \times 3$ & 1 & 3\\
        15 & $32 \times 192 \times 1 \times 1$ & 96 & 1\\
        16 & $192 \times 32 \times 1 \times 1$ & 16 & 1\\
        17 & $192 \times 1 \times 3 \times 3$ & 1 & 2\\
        18 & $32 \times 192 \times 1 \times 1$ & 96 & 1\\
        19 & $192 \times 32 \times 1 \times 1$ & 16 & 1\\
        20 & $192 \times 1 \times 3 \times 3$ & 1 & 2\\
        21 & $64 \times 192 \times 1 \times 1$ & 96 & 1\\
        22 & $384 \times 64 \times 1 \times 1$ & 32 & 1\\
        23 & $384 \times 1 \times 3 \times 3$ & 1 & 2\\
        24 & $64 \times 384 \times 1 \times 1$ & 192 & 1\\
        25 & $384 \times 64 \times 1 \times 1$ & 32 & 1\\
        26 & $384 \times 1 \times 3 \times 3$ & 1 & 2\\
        27 & $64 \times 384 \times 1 \times 1$ & 192 & 1\\
        28 & $384 \times 64 \times 1 \times 1$ & 32 & 1\\
        29 & $384 \times 1 \times 3 \times 3$ & 1 & 2\\
        30 & $64 \times 384 \times 1 \times 1$ & 192 & 1\\
        31 & $384 \times 64 \times 1 \times 1$ & 32 & 1\\
        32 & $384 \times 1 \times 3 \times 3$ & 1 & 2\\
        33 & $96 \times 384 \times 1 \times 1$ & 192 & 1\\
        34 & $576 \times 96 \times 1 \times 1$ & 48 & 1\\
        35 & $576 \times 1 \times 3 \times 3$ & 1 & 2\\
        36 & $96 \times 576 \times 1 \times 1$ & 288 & 1\\
        37 & $576 \times 96 \times 1 \times 1$ & 48 & 1\\
        38 & $576 \times 1 \times 3 \times 3$ & 1 & 2\\
        39 & $96 \times 576 \times 1 \times 1$ & 288 & 1\\
        40 & $576 \times 96 \times 1 \times 1$ & 48 & 1\\
        41 & $576 \times 1 \times 3 \times 3$ & 1 & 2\\
        42 & $160 \times 576 \times 1 \times 1$ & 288 & 1\\
        43 & $960 \times 160 \times 1 \times 1$ & 80 & 1\\
        44 & $960 \times 1 \times 3 \times 3$ & 1 & 2\\
        45 & $160 \times 960 \times 1 \times 1$ & 480 & 1\\
        46 & $960 \times 160 \times 1 \times 1$ & 80 & 1\\
        47 & $960 \times 1 \times 3 \times 3$ & 1 & 2\\
        48 & $160 \times 960 \times 1 \times 1$ & 480 & 1\\
        49 & $960 \times 160 \times 1 \times 1$ & 80 & 1\\
        50 & $960 \times 1 \times 3 \times 3$ & 1 & 3\\
        51 & $320 \times 960 \times 1 \times 1$ & 480 & 1\\
        52 & $1280 \times 320 \times 1 \times 1$ & 160 & 1\\
        \scriptsize{classifier} & $1000 \times 1280 \times 1 \times 1$ & 560 & 1\\
        \bottomrule 
    \end{tabular}
    \label{table:struct-mv2-b}
    \end{minipage}
\end{table}

\begin{table}[h]
    \scriptsize
    \centering
    \caption{Layerwise $\{c,n\}$ configuration for Struct-EffNet architecture}
    \begin{tabular}{cccc}
        \toprule
        Idx & \shortstack{Dimension\\$C_{out}\times C\times N\times N$} & $c$ & $n$ \\
        % % \cmidrule(lr){2-4}
        % & $C_{out}$ & $C$ & $N$ & \\
        \midrule
        1 & $32 \times 3 \times 3 \times 3$ & 3 & 3\\
        2 & $32 \times 1 \times 3 \times 3$ & 1 & 3\\
        3 & $16 \times 32 \times 1 \times 1$ & 32 & 1\\
        4 & $16 \times 1 \times 3 \times 3$ & 1 & 3\\
        5 & $16 \times 16 \times 1 \times 1$ & 16 & 1\\
        6 & $96 \times 16 \times 1 \times 1$ & 16 & 1\\
        7 & $96 \times 1 \times 3 \times 3$ & 1 & 3\\
        8 & $24 \times 96 \times 1 \times 1$ & 96 & 1\\
        9 & $144 \times 24 \times 1 \times 1$ & 24 & 1\\
        10 & $144 \times 1 \times 3 \times 3$ & 1 & 3\\
        11 & $24 \times 144 \times 1 \times 1$ & 144 & 1\\
        12 & $144 \times 24 \times 1 \times 1$ & 24 & 1\\
        13 & $144 \times 1 \times 3 \times 3$ & 1 & 3\\
        14 & $24 \times 144 \times 1 \times 1$ & 144 & 1\\
        15 & $144 \times 24 \times 1 \times 1$ & 24 & 1\\
        16 & $144 \times 1 \times 5 \times 5$ & 1 & 5\\
        17 & $40 \times 144 \times 1 \times 1$ & 144 & 1\\
        18 & $240 \times 40 \times 1 \times 1$ & 40 & 1\\
        19 & $240 \times 1 \times 5 \times 5$ & 1 & 5\\
        20 & $40 \times 240 \times 1 \times 1$ & 240 & 1\\
        21 & $240 \times 40 \times 1 \times 1$ & 40 & 1\\
        22 & $240 \times 1 \times 5 \times 5$ & 1 & 5\\
        23 & $40 \times 240 \times 1 \times 1$ & 240 & 1\\
        24 & $240 \times 40 \times 1 \times 1$ & 40 & 1\\
        25 & $240 \times 1 \times 3 \times 3$ & 1 & 3\\
        26 & $80 \times 240 \times 1 \times 1$ & 240 & 1\\
        27 & $480 \times 80 \times 1 \times 1$ & 64 & 1\\
        28 & $480 \times 1 \times 3 \times 3$ & 1 & 3\\
        29 & $80 \times 480 \times 1 \times 1$ & 360 & 1\\
        30 & $480 \times 80 \times 1 \times 1$ & 64 & 1\\
        31 & $480 \times 1 \times 3 \times 3$ & 1 & 3\\
        32 & $80 \times 480 \times 1 \times 1$ & 360 & 1\\
        33 & $480 \times 80 \times 1 \times 1$ & 64 & 1\\
        34 & $480 \times 1 \times 3 \times 3$ & 1 & 3\\
        35 & $80 \times 480 \times 1 \times 1$ & 360 & 1\\
        36 & $480 \times 80 \times 1 \times 1$ & 64 & 1\\
        37 & $480 \times 1 \times 5 \times 5$ & 1 & 5\\
        38 & $112 \times 480 \times 1 \times 1$ & 360 & 1\\
        39 & $672 \times 112 \times 1 \times 1$ & 80 & 1\\
        40 & $672 \times 1 \times 5 \times 5$ & 1 & 5\\
        41 & $112 \times 672 \times 1 \times 1$ & 560 & 1\\
        42 & $672 \times 112 \times 1 \times 1$ & 96 & 1\\
        43 & $672 \times 1 \times 5 \times 5$ & 1 & 5\\
        44 & $112 \times 672 \times 1 \times 1$ & 560 & 1\\
        45 & $672 \times 112 \times 1 \times 1$ & 96 & 1\\
        46 & $672 \times 1 \times 5 \times 5$ & 1 & 5\\
        47 & $112 \times 672 \times 1 \times 1$ & 560 & 1\\
        48 & $672 \times 112 \times 1 \times 1$ & 96 & 1\\
        49 & $672 \times 1 \times 5 \times 5$ & 1 & 5\\
        50 & $192 \times 672 \times 1 \times 1$ & 560 & 1\\
        51 & $1152 \times 192 \times 1 \times 1$ & 100 & 1\\
        52 & $1152 \times 1 \times 5 \times 5$ & 1 & 5\\
        53 & $192 \times 1152 \times 1 \times 1$ & 640 & 1\\
        54 & $1152 \times 192 \times 1 \times 1$ & 100 & 1\\
        55 & $1152 \times 1 \times 5 \times 5$ & 1 & 5\\
        56 & $192 \times 1152 \times 1 \times 1$ & 640 & 1\\
        57 & $1152 \times 192 \times 1 \times 1$ & 100 & 1\\
        58 & $1152 \times 1 \times 5 \times 5$ & 1 & 5\\
        59 & $192 \times 1152 \times 1 \times 1$ & 640 & 1\\
        60 & $1152 \times 192 \times 1 \times 1$ & 100 & 1\\
        61 & $1152 \times 1 \times 5 \times 5$ & 1 & 5\\
        62 & $192 \times 1152 \times 1 \times 1$ & 576 & 1\\
        63 & $1152 \times 192 \times 1 \times 1$ & 160 & 1\\
        64 & $1152 \times 1 \times 3 \times 3$ & 1 & 3\\
        65 & $320 \times 1152 \times 1 \times 1$ & 576 & 1\\
        66 & $1920 \times 320 \times 1 \times 1$ & 160 & 1\\
        67 & $1920 \times 1 \times 3 \times 3$ & 1 & 3\\
        68 & $320 \times 1920 \times 1 \times 1$ & 960 & 1\\
        69 & $1280 \times 320 \times 1 \times 1$ & 160 & 1\\
        \scriptsize{classifier} & $1000 \times 1280 \times 1 \times 1$ & 480 & 1\\
        \bottomrule 
    \end{tabular}
    \label{table:struct-effnet}
\end{table}

\end{document}